\documentclass[10pt,twocolumn,letterpaper]{article}

\usepackage[pagenumbers]{cvpr} 

\usepackage[dvipsnames]{xcolor}

\definecolor{cvprblue}{rgb}{0.21,0.49,0.74}
\usepackage[pagebackref,breaklinks,colorlinks,citecolor=cvprblue]{hyperref}
\usepackage{graphicx}
\usepackage{bm}
\usepackage{booktabs}
\usepackage{multirow}
\usepackage{makecell}
\usepackage[most]{tcolorbox}
\usepackage{amsfonts}
\usepackage{nicefrac}
\usepackage{microtype}
\usepackage{xcolor}
\usepackage{amsmath}
\usepackage{textcomp}
\usepackage{placeins} 
\usepackage{listings}
\tcbuselibrary{breakable} 

\usepackage{algorithm,algpseudocode}
\usepackage{threeparttable}
\definecolor{mydarkblue}{rgb}{0,0.08,1}
\definecolor{mydarkred}{rgb}{0.8,0.02,0.02}
\definecolor{mydarkorange}{rgb}{0.40,0.2,0.02}
\definecolor{mypurple}{RGB}{111,0,255}
\definecolor{myred}{rgb}{1.0,0.0,0.0}
\definecolor{mygold}{rgb}{0.75,0.6,0.12}
\definecolor{mydarkgray}{rgb}{0.66, 0.66, 0.66}
\definecolor{mydarkgreen}{rgb}{0.02,0.6,0.02}

\newcommand{\rootname}{\textsc{Root}\xspace}
\newcommand{\rootlogo}[1][1.85em]{%
  \raisebox{-0.3\height}{\includegraphics[height=#1]{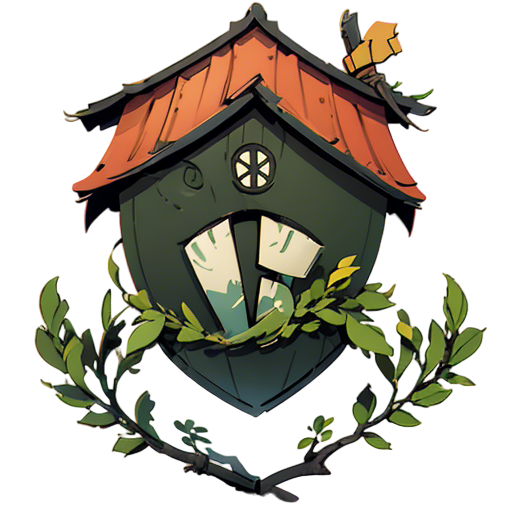}}
}

\title{\rootlogo\rootname: VLM based System for Indoor Scene Understanding and Beyond}

\author{
    \textbf{Yonghui Wang}$^{1,2}$\thanks{Work done during internship at Tencent},
    \textbf{Shi-Yong Chen}$^{2}$,
    \textbf{Zhenxing Zhou}$^{2}$,
    \textbf{Siyi Li}$^{2}$,
    \textbf{Haoran Li}$^{1,2}$, \\
    \textbf{Wengang Zhou}$^{1}$,
    \textbf{Houqiang Li}$^{1}$ \\[2pt]
    $^1$University of Science and Technology of China;
    $^2$Game AI Center, Tencent IEG \\[2pt]
    \texttt{wyh1998@mail.ustc.edu.cn}
}

\begin{document}

\maketitle
\begin{abstract}
Recently, Vision Language Models (VLMs) have experienced significant advancements, yet these models still face challenges in spatial hierarchical reasoning within indoor scenes.
In this study, we introduce \rootname\footnote{Our system, inspired by the \textbf{Roo}m of Requiremen\textbf{t} (\rootname) from Harry Potter—known for adapting to users' needs—helps people understand indoor scenes, which can vary in countless ways.}, a VLM-based system designed to enhance the analysis of indoor scenes.
Specifically, we first develop an iterative object perception algorithm using GPT-4V to detect object entities within indoor scenes.
This is followed by employing vision foundation models to acquire additional meta-information about the scene, such as bounding boxes.
Building on this foundational data, we propose a specialized VLM, SceneVLM, which is capable of generating spatial hierarchical scene graphs and providing distance information for objects within indoor environments.
This information enhances our understanding of the spatial arrangement of indoor scenes.
To train our SceneVLM, we collect over 610,000 images from various public indoor datasets and implement a scene data generation pipeline with a semi-automated technique to establish relationships and estimate distances among indoor objects. 
By utilizing this enriched data, we conduct various training recipes and finish SceneVLM.
Our experiments demonstrate that \rootname facilitates indoor scene understanding and proves effective in diverse downstream applications, such as 3D scene generation and embodied AI.
The code will be released at \url{https://github.com/harrytea/ROOT}.
\end{abstract}
    
\section{Introduction}
\label{sec:intro}

\begin{figure}[tbp]
\begin{center}
\includegraphics[width=0.98\linewidth]{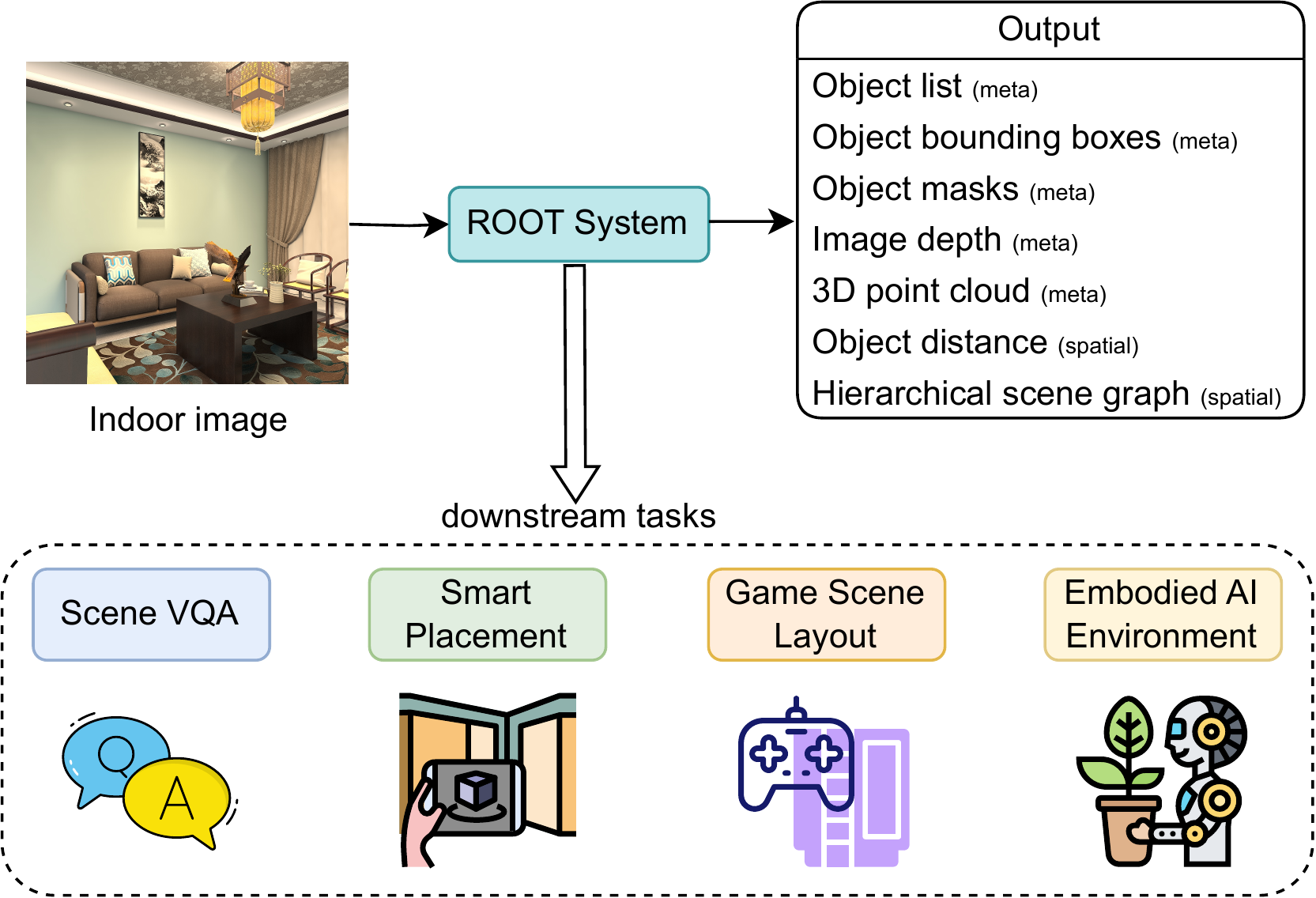}
\end{center}
\vspace{-0.1in}
\caption{\rootname is a system designed to interpret indoor scene images and extract various types of meta-information about the scenes. Utilizing this information, \rootname can generate hierarchical relationships and spatial distances among indoor objects. This enriched data serves to support various downstream tasks.}
\label{fig:intro}
\end{figure}

Indoor scene understanding is a critical task and has been extensively studied~\cite{cordts2016cityscapes,balazevic2024towards,peng2023openscene,xiao2018unified,zhang2017deepcontext}.
The advent of VLMs has notably advanced this field, demonstrating their robust zero-shot learning capabilities~\cite{cao2024maplm,fu2024scene,zhang2024agent3d,ha2022semantic}. 
This task encompasses a myriad of information, such as the entities within a room, their positions, and their interrelationships. 
This information is essential for excelling in various downstream tasks, including intelligent object placement~\cite{yi2022human}, 3D scene generation~\cite{yang2024physcene,yang2024holodeck}, and improving the performance of domestic robots in executing human commands~\cite{zhao2021luminous,wang2024embodiedscan}.
However, a notable challenge in indoor scene understanding is the comprehension of spatial relationships, particularly the limited perception of these relationships by VLMs.

Most general-purpose VLMs are trained with a substantial volume of high-quality instruction-following data, enabling them to comprehend image content and perform standard tasks such as Visual Question Answering (VQA)~\cite{bai2023qwen,wang2024qwen2,chen2024internvl,liu2024visual}. 
However, these models face significant challenges in parsing indoor scenes, a critical barrier in the pursuit of Artificial General Intelligence (AGI).
We argue that the ability to understand indoor scenes is an essential aspect of VLMs, as it supports the advancement of various downstream tasks~\cite{chen2024spatialvlm}.
This paper primarily focuses on the understanding of indoor scenes, especially in terms of spatial perception.
As depicted in Figure~\ref{fig:intro}, we introduce \rootname, a VLM-based system designed to interpret indoor scenes by identifying objects and their attributes, and ultimately determining the hierarchical positional relationships and distance information among these objects.
This enhanced understanding facilitates the development of new techniques to improve performance in downstream tasks, such as scene-based VQA and intelligent object placement.

To achieve our objectives, we employ a variety of readily available foundation models and custom models to analyze indoor scenes, culminating in the creation of our system, \rootname. 
Our process is divided into three parts: \emph{iterative object perception, indoor scene parsing, and hierarchical scene graph generation}.
Initially, we utilize a GPT-4V~\cite{2023GPT4VisionSC} based method for perceiving indoor objects to identify entities within the scene. 
To detect smaller objects, we adopt an iterative approach that involves magnifying and re-detecting specific areas as necessary.
Subsequently, we use existing vision foundation models to parse indoor scenes, extracting depth information and basic object attributes such as bounding boxes and masks.
Finally, our customized model, SceneVLM, utilizes the data from the preceding steps to generate a hierarchical scene graph of the indoor objects along with spatial distance information.

To train SceneVLM, we develop a scene data generation pipeline that semi-automatically produces training data with human assistance.
To ensure robust zero-shot capabilities for the model, we gather a diverse dataset of over 610,000 indoor scene images. 
We then employ the CLIP model~\cite{fang2023data} to filter out unsuitable images.
Leveraging the capabilities developed in the initial steps, we automate the generation of distance data and semi-automatically construct the hierarchical data between objects.
Using the data generated by our pipeline, we conduct experiments on advanced open-source VLM models to enhance their spatial understanding of indoor environments.

In conclusion, our \rootname system exhibits the following capabilities.
First, it processes an RGB image of an indoor scene to identify objects and analyze their attributes as well as those of the scene. 
Moreover, it models the spatial relationships among these objects, generating a scene graph that delineates the hierarchical relationships and distances between them.

Our contributions are summarized as follows:
\begin{itemize} 
\item We introduce \rootname, a VLM based system designed for indoor scene understanding, capable of extracting meta-information from images and delineating the hierarchical spatial relationships among objects. 
\item We develop a scene data generation pipeline to create a spatial scene dataset and introduce SceneVLM to aggregate existing attribute information of objects within rooms, thereby generating spatial information for indoor scenes. We explore various training recipes to evaluate their impact on the performance of SceneVLM. 
\item We effectively demonstrate the significant applications of our method in specific downstream tasks, which enable further advancements that contribute to enhanced performance in these areas.
\end{itemize}

\section{Related work}
\label{sec:related}
\noindent
\textbf{Indoor Scene Understanding.}
Scene understanding is a fundamental task in computer vision, broadly encompassing various sub-tasks such as scene segmentation~\cite{yu2020context,fu2019dual,porzi2019seamless}, depth estimation~\cite{saxena2023monocular,yang2024depth}, room layout analysis~\cite{shen2023disentangling,tsai2024no}, 3D reconstruction~\cite{chen2024polydiffuse}, and dynamic scenes~\cite{hu2020probabilistic,tosi2020distilled}.
Recent advancements in robotics~\cite{li2021robotic} and mixed reality~\cite{keshavarzi2022mutual} underscore the significance of understanding indoor scenes, positioning it as a dynamic research area aimed at enhancing model generalization through increased data diversity and richness~\cite{dong2023shape,yang2024swin3d}.
OpenScene~\cite{peng2023openscene} incorporates 3D scene features into the textual and visual spaces of CLIP~\cite{radford2021learning}.
Similarly, PLA~\cite{ding2023pla} leverages the knowledge embedded in pre-trained vision-language foundation models by associating 3D features with semantically rich captions, thereby facilitating open-vocabulary understanding of indoor scenes.

Despite the innovative methods employed in numerous scene analysis techniques, they often struggle with generalizing to new scenes due to the limitations of manually crafted rules and the diversity of training datasets.
The robust zero-shot capabilities of VLMs introduce new avenues for advancing scene understanding. 
Our research concentrates on indoor scenes depicted in RGB images and introduces a new system for indoor scene understanding.
Normally, the human visual system excels at perceiving local visual details and performing semantic and geometric reasoning to comprehend complex object relationships. 
Similarly, our system aims to comprehend scenes, including both attribute and hierarchical relationships among objects, which can help the AI agent to effectively respond to human commands.

\begin{figure*}[tbp]
\begin{center}
\includegraphics[width=1.0\linewidth]{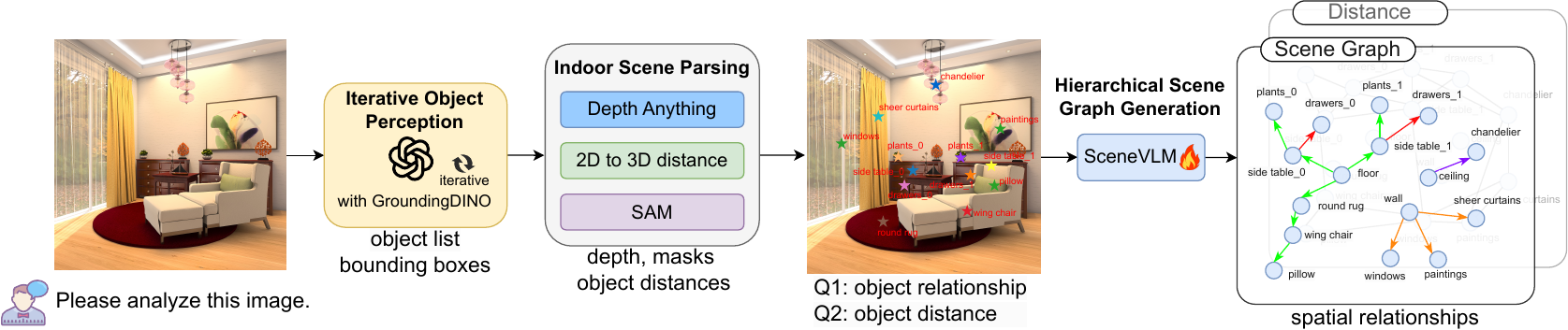}
\end{center}
\vspace{-0.1in}
\caption{We introduce \rootname, a system designed for understanding indoor scenes. Initially, the system utilizes an iterative object perception module based on GPT-4V to identify entities within a given image. Subsequently, the indoor scene and objects are parsed using existing vision foundation models to gather meta-information about the scene. Finally, the object information is processed by SceneVLM, resulting in a scene graph that illustrates the spatial hierarchical relationships and distance information. In the scene graph, arrows of different colors denote different relationships.}
\label{fig:arch}
\end{figure*}

\noindent
\textbf{Spatial Reasoning in Vision Language Models.}
VLMs exhibit proficiency in general visual tasks such as image captioning and VQA.
However, they face challenges in spatial reasoning, particularly in tasks requiring precise visual direction and localization.
To mitigate these challenges, SpatialVLM\cite{chen2024spatialvlm} enhances the spatial reasoning capabilities of VLMs through automated data generation and specialized training techniques aimed at distance estimation.
This advancement enables VLMs to perceive distances between objects and handle complex spatial reasoning tasks. 
Furthermore, TopViewRS\cite{li2024topviewrs} explores VLMs from a top-view perspective, evaluating their comprehension of top views and spatial relationships, and highlighting the difficulties they face in understanding spatial layouts from such perspectives.
Moreover, Wang \emph{et al.}~\cite{wang2024picture} also notes that VLMs often struggle with spatial reasoning, likely due to the simplistic processing of visual signals in existing VLM architectures. 

Current research on models predicting spatial relationships in indoor scene arrangements is less explored. In this paper, we elucidate the hierarchical relationships and spatial distance among objects in indoor scenes, allowing VLMs to directly learn the implicit representations between objects from RGB images and to model their spatial relevance.

\noindent
\textbf{Scene Graph Generation.}
Scene Graph Generation (SGG) aims to transform visual scenes into explicit graphical representations, explicitly delineating objects and their interrelationships within a scene.
SGG aids in structuring and interpreting visual scenes by forming subject-relation-object triplets among objects in an image and has been widely applied in various vision-language tasks, such as visual question answering\cite{hildebrandt2020scene,qian2022scene}, image description\cite{zhong2020comprehensive}, referring expressions\cite{yang2020graph}, and image retrieval\cite{schroeder2020structured}. 
Recent studies have begun to utilize the image-text matching capabilities of pretrained VLMs to tackle multiple SGG challenges in open vocabulary settings~\cite{li2024pixels,wang2024llava}. 

In this paper, we focus on constructing indoor object scene graphs in open vocabulary settings. 
To achieve this, we define four types of hierarchical relationships for indoor objects, employ open-source vision foundation models to parse objects in images, and input them into VLMs. 
This process allows VLMs to generate structured scene graph information delineating the relationships between objects from RGB images.	
\section{\rootname}
As shown in Figure~\ref{fig:arch}, our \rootname system consists of three main components: iterative object perception, indoor scene parsing, and hierarchical scene graph generation.
The first component identifies objects within the indoor scene.
Then, the second gathers meta-information about the objects and the scene. 
Finally, the third utilizes this information to generate a hierarchical scene graph and estimate distance.
Leveraging various foundation models, our system demonstrates superior performance in understanding indoor scenes.

\subsection{Iterative Object Perception}
We employ GPT-4V to identify objects within indoor environments, leveraging its exceptional multimodal capabilities.
GPT-4V allows for a deep understanding of object semantics.
Its robust zero-shot capabilities enable it to perform effectively in novel environments by drawing on its extensive world knowledge.
Upon processing an indoor image $I_{in}$, GPT-4V is prompted to generate a list of objects $\{o_{i}\}_{i=1}^{N}$, where $i$ denotes the $i_{th}$ object, along with an indication of whether each object qualifies as a container $\{c_{i}\}_{i=1}^{N}$ (true or false).
Subsequently, we employ GroundingDINO~\cite{liu2023grounding} to detect these objects and produce bounding boxes $\{b_{ij}\}_{i=1,j=1}^{N,M}$, where $i$ is the $i_{th}$ object and $j$ is the candidate bounding box for the $i_{th}$ object.
We assess the output of GroundingDINO, retaining bounding boxes with probabilities exceeding $p_{m}$.
If all bounding boxes fall below this threshold, the object is discarded.
When multiple bounding boxes are candidates for a single object, we compare the scores of the top two; if their difference exceeds $p_{n}$, we select the larger bounding box as definitive.
Conversely, if the difference is less than $p_{n}$, GPT-4V is prompted to determine the most suitable bounding box for further analysis.
For objects identified as containers, we increase their bounding box dimensions by a factor of $S$ and crop them to restart the detection processes and update the object list.
This iterative refinement process ensures that GPT-4V accurately identifies as many objects as possible, which is particularly beneficial for the perception of small objects, thereby enhancing the precision of object detection and augmenting the system's ability to process and interpret complex indoor scenes.
Details of the complete algorithm are provided in the supplementary material.

\subsection{Indoor Scene Parsing}
The essential attributes of indoor scenes encompass objects, bounding boxes, masks, and point cloud data.
In this study, we parse the indoor scenes to extract detailed information.
Initially, objects and their bounding boxes are acquired.
These bounding boxes are then utilized to prompt SAM~\cite{kirillov2023segment}, which generates a mask for each identified object.
Concurrently, the DepthAnything~\cite{yang2024depth} model processes the original image, enabling the extraction of depth information and the generation of a three-dimensional (3D) point cloud representing the current indoor environment.
By integrating the 3D point cloud data with the mask information, we derive the 3D point cloud representation for each object.
Subsequently, the spatial distance between objects is determined by calculating the centroid distances of their respective point clouds.
This metric provides insights into the spatial arrangement and proximity of objects within the scene. 
Through these steps, we successfully obtain comprehensive meta-information for each object in the original image, facilitating a deeper understanding of the indoor environment.

\begin{figure}[tbp]
\begin{center}
\includegraphics[width=0.95\linewidth]{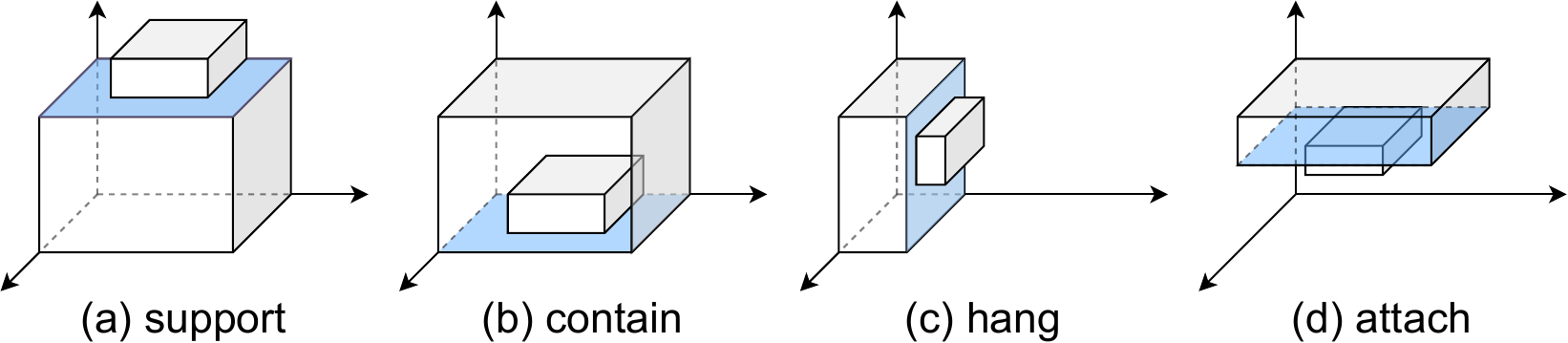}
\end{center}
\vspace{-0.1in}
\caption{Four types of hierarchical relationships as defined. In each sub-figure, the larger object represents the parent object, while the smaller object denotes the child object.}
\label{fig:relation}
\end{figure}

\begin{figure}[tbp]
\begin{center}
\includegraphics[width=0.95\linewidth]{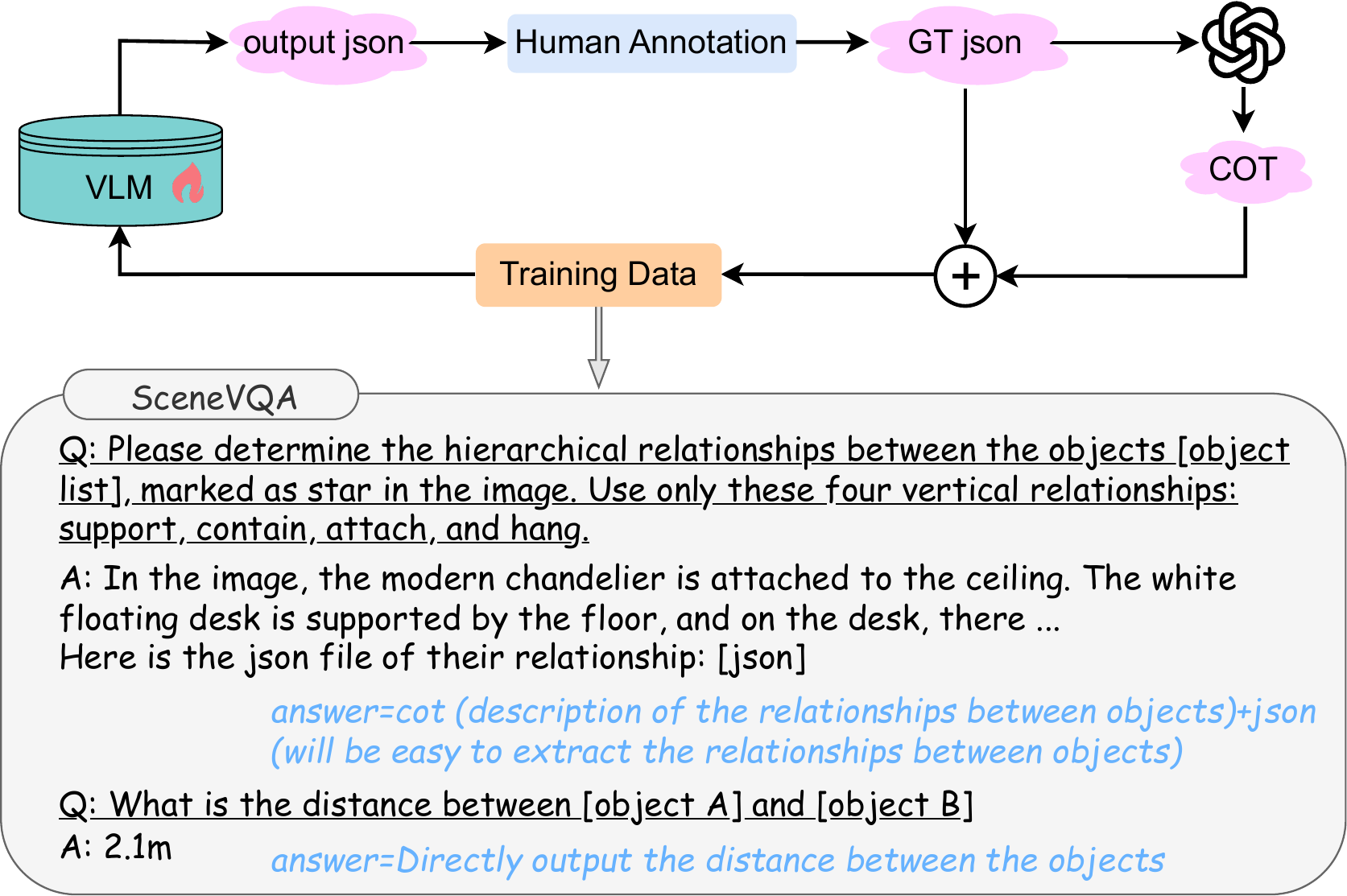}
\end{center}
\vspace{-0.1in}
\caption{SceneVQA data generation pipeline. This diagram depicts the semi-automated pipeline used to create GraphVQA data, which includes manual annotation, GPT-4 assisted transformation, and iterative refinement. For DistanceVQA, object distances are computed directly from 3D point cloud data.}
\label{fig:vlm_pipeline}
\end{figure}

\subsection{Hierarchical Scene Graph Generation}
Our objective in this step is to generate hierarchical scene graphs and spatial distances of objects in indoor environments. 
To achieve this, we collect data using the established pipeline, resulting in the SceneVQA dataset. 
Using this dataset, we develop a specialized model named SceneVLM. 
The following sections detail this process.

\noindent
\textbf{Data collection.}
We curate training data from a variety of open-source scene datasets, including 3D-Future~\cite{fu20213d}, TUM~\cite{sturm2012benchmark}, SUN~\cite{xiao2016sun}, MIT Indoor Scenes~\cite{quattoni2009recognizing}, and Places~\cite{zhou2017places}, which collectively provide a diverse array of scenes. 
Notably, the TUM dataset contains video data, from which we randomly sample frames to extract images of indoor scenes. 
For the other datasets, we selectively retain only the data relevant to indoor environments, discarding any outdoor scenes. 
Furthermore, to ensure the quality of the indoor scene data, we employ a CLIP model~\cite{fang2023data} to rigorously evaluate and filter the input images, specifically excluding those of low quality or those that depict only fragments or isolated objects of indoor settings. 
For more details, please refer to the supplementary material.

\begin{table*}[t]
    \centering
    \small
    \resizebox{1.0\linewidth}{!}{%
        \begin{tabular}{llcccccccccccccccccc}
            \Xhline{2.5\arrayrulewidth}
            \multirow{3}{*}{Method} & \multirow{3}{*}{LLM} & \multirow{3}{*}{JSON \%} & \multicolumn{4}{c}{Pairwise Relation Accuracy} & \multicolumn{4}{c}{Object-wise Relation Accuracy} & \multicolumn{4}{c}{Layer-wise Accuracy} & \multicolumn{4}{c}{Node Detection Accuracy} \\
            \cmidrule(lr){4-7} \cmidrule(lr){8-11} \cmidrule(lr){12-15} \cmidrule(lr){16-19}
            & & & Precision & Recall & F-score & IoU & Precision & Recall & F-score & IoU & Precision & Recall & F-score & IoU & Precision & Recall & F-score & IoU \\
            \hline
            InstructBLIP~\cite{dai2023instruct} & Vicuna-13B & 1.22 & 0 & 0 & 0 & 0 & 0.01 & 0.01 & 0.01 & 0 & 0.92 & 0.28 & 0.40 & 0.26 & 0.92 & 0.28 & 0.40 & 0.26 \\
            LLaVA-1.5~\cite{liu2024improved} & Vicuna-13B & 0 & - & - & - & - & - & - & - & - & - & - & - & - & - & - & - & - \\
            LLaVA-NeXT~\cite{liu2024llavanext} & Vicuna-13B & 88.11 & 0.26 & 0.21 & 0.23 & 0.13 & 0.60 & 0.53 & 0.54 & 0.34 & 21.49 & 35.23 & 25.92 & 15.38 & 65.97 & 34.32 & 42.93 & 29.71 \\
            Qwen2-VL~\cite{wang2024qwen2} & Qwen2-7B & 65.00 & 5.01 & 5.17 & 4.81 & 2.98 & 6.48 & 6.32 & 6.28 & 4.07 & 13.29 & 24.56 & 16.66 & 10.03 & 50.06 & 39.58 & 42.48 & 33.72 \\
            InternVL2~\cite{chen2024internvl} & InternLM2.5-7B & 83.65 & 0.89 & 0.87 & 0.86 & 0.52 & 6.29 & 4.63 & 5.04 & 3.02 & 61.85 & 28.98 & 36.25 & 25.35 & 79.05 & 36.98 & 46.47 & 34.15 \\
            MiniCPM-V 2.6~\cite{yao2024minicpm} & Qwen2-7B & 87.16 & 2.75 & 3.84 & 2.87 & 1.60 & 5.06 & 4.49 & 4.58 & 2.64 & 37.76 & 27.53 & 27.08 & 18.25 & 74.54 & 45.70 & 52.49 & 40.40 \\
            LlaMA-3.2~\cite{dubey2024llama} & LlaMA3.1-8B & 81.89 & 1.06 & 1.31 & 1.09 & 0.63 & 3.54 & 2.69 & 2.92 & 1.72 & 39.48 & 30.00 & 29.31 & 19.65 & 68.64 & 34.02 & 42.79 & 30.82 \\
            GLM-4V~\cite{glm2024chatglm} & GLM-4-9B & 97.57 & 4.09 & 3.73 & 3.52 & 2.09 & 5.28 & 4.31 & 4.54 & 2.73 & 70.13 & 38.45 & 47.62 & 33.08 & 92.39 & 38.91 & 51.34 & 36.81 \\
            Gemini Pro & - & 94.73 & 3.59 & 3.80 & 3.45 & 1.96 & 7.16 & 4.84 & 5.52 & 3.31 & 53.35 & 38.72 & 39.35 & 25.99 & 80.55 & 38.05 & 48.59 & 34.27 \\
            GPT-4V & - & \underline{98.78} & 48.61 & 48.60 & 48.00 & 37.68 & 48.23 & 47.92 & 47.83 & 36.77 & 62.36 & 52.09 & 52.72 & 40.14 & 94.71 & 81.13 & 84.24 & 77.98 \\
            \hline
            SceneVLM & Qwen2-7B & \textbf{100} & \textbf{91.37} & \underline{90.38} & \underline{90.76} & \underline{85.02} & \underline{87.68} & \underline{87.17} & \underline{87.39} & \underline{80.22} & \textbf{77.93} & \textbf{78.53} & \textbf{77.96} & \textbf{72.52} & \textbf{99.97} & \textbf{99.30} & \textbf{99.58} & \textbf{99.28} \\
            SceneVLM & InternLM2.5-7B & \textbf{100} & \underline{91.33} & \textbf{90.62} & \textbf{90.85} & \textbf{85.24} & \textbf{88.04} & \textbf{87.41} & \textbf{87.68} & \textbf{80.77} & \underline{77.89} & \underline{77.70} & \underline{77.55} & \underline{71.99} & \underline{99.60} & \underline{98.76} & \underline{99.11} & \underline{98.55} \\
            \Xhline{2.5\arrayrulewidth}
        \end{tabular}
    }
    \caption{Quantitative comparison results of our method with other VLMs across four perspectives using the metrics Precision, Recall, F-score, and IoU. ``JSON'' indicates the percentage of JSON files generated accurately for loading. The best and the second results are highlighted in \textbf{blod} and \underline{underlined}, respectively. The metrics are scaled by a factor of 100 for enhanced clarity.}
    \label{tab:sgg}
\end{table*}

\begin{table}[t]
\centering
\resizebox{\linewidth}{!}{
    \begin{tabular}{lccccc}
        \Xhline{2.5\arrayrulewidth}
        & 3D-FUTURE & TUM & SUN  &  MIT Indoor & Places  \\
        \hline
        Graph data       & 1322 & 53 & 512 & 98 & 7776 \\
        Distance data       & 12774 & 437 & 5368 & 4443 & 595839 \\
        Test data        & 80 & 20 & 40 & 40 & 560 \\
        \hline
        \multicolumn{6}{c}{Total object categories: 322,064; Total scenes: over 40} \\
        \Xhline{2.5\arrayrulewidth}
    \end{tabular}
}
\caption{Data statistics of our created SceneVQA dataset.}
\label{tab:data_stat}
\end{table}

\noindent
\textbf{Hierarchical relationship defination.}
Most research focuses on predicting floor plans in indoor environments. 
However, we argue that recognizing hierarchical relationships between objects in indoor scenes is crucial for VLMs to comprehend scene layouts. 
In this paper, we define four types of hierarchical relationships: support, contain, hang, and attach. 
These relationships facilitate a deeper understanding of the underlying logic governing object arrangement in indoor settings.
As shown in Figure~\ref{fig:relation}, each sub-figure demonstrates these relationships, with the larger block representing the parent object and the smaller block representing the child object.
``support'' indicates that the child object is supported by the upper surface of the parent. 
``contain'' means that the child object is enclosed within the internal space of the parent object.
``hang'' denotes that the child is suspended from the parent object, while ``attach'' suggests that the child object is positioned below the parent object.

\noindent
\textbf{SceneVQA dataset.}
We develop the SceneVQA dataset, which comprises two components: GraphVQA and DistanceVQA.
GraphVQA focuses on hierarchical relationships among objects, while DistanceVQA emphasizes the spatial distances between them.
Typically, LLMs are pretrained on extensive datasets containing world knowledge, which implicitly include information about the arrangement of indoor objects, such as a bed supporting a pillow.
However, real-world indoor scenes are often more complex due to human activities, leading to situations where the pillow may fall to the floor.
Therefore, the model must truly ``understand'' the content in the image to make accurate predictions. 

To generate the GraphVQA data, we implement a semi-automated annotation process.
As shown in Figure~\ref{fig:vlm_pipeline}, we begin by manually annotating hierarchical scene graphs to serve as ground-truth data in JSON format.
Given the challenge of VLM directly outputting hierarchical JSON relationships, we employ GPT-4~\cite{achiam2023gpt} to transform manually annotated JSON files into natural language descriptions serving as chain-of-thought data. 
We find that translating from natural language to JSON is more straightforward than direct JSON generation.
We retrain the VLM with these new descriptions and JSON content, iteratively refining the VQA data.
With the pre-annotated data, we train the VLM, which subsequently generates new data.
These outputs are manually reviewed and corrected by human annotators to rectify any inaccuracies or omissions in object relationships, ensuring the data's accuracy and reliability through this iterative process.
For DistanceVQA, we utilize 3D point cloud data to develop a dataset that provides distances between objects. 
Rather than using positional terms like ``in front of'' or ``behind'' we opt for a direct distance representation to describe the distances between two objects, \emph{e.g.}, 2.1m.
Together, these two VQA datasets form SceneVQA, playing a crucial role in training and refining the SceneVLM, thereby enhancing its ability to understand indoor scenes.
Note, before the creation of SceneVQA, these indoor images must undergo the first two steps to generate objects for annotation by annotators.

\noindent
\textbf{Dataset statistics.}
As shown in Table~\ref{tab:data_stat}, our dataset comprises over 40 types of indoor scenes, each containing an average of 15.4 objects.
For GraphVQA, due to the complexity of the semi-automated process (with an annotation time of approximately 4-5 minutes per image), and the inherent prior knowledge of VLMs which facilitates easier learning, we annotated 9,761 images. 
For DistanceVQA, leveraging a fully automated process and the challenges VLMs face in understanding spatial relationships, over 610,000 entries are collected. 
Additionally, with the aid of GPT-4V, we identify more than 320,000 categories featuring various adjectives of the same type within these images.
\section{Experiments}

\begin{table}[t]
\centering
\resizebox{\linewidth}{!}{
    \begin{tabular}{llccc}
        \Xhline{2.5\arrayrulewidth}
        Method & LLM & Number \% & Range [80,120] & Range [50,200]   \\
        \hline
        InstructBLIP~\cite{dai2023instruct} &  Vicuna-13B & 0.01 & 0.01 & 0.01 \\
        LLaVA-1.5~\cite{liu2024improved} & Vicuna-13B & 94.7 & 19.26 & 55.56 \\
        LLaVA-NeXT~\cite{liu2024llavanext} & Vicuna-13B & 61.27 & 7.71 & 24.31 \\
        GLM-4V~\cite{glm2024chatglm} & GLM-4-9B & 94.54 & 12.36 & 39.07 \\
        LlaMA-3.2~\cite{dubey2024llama} & LlaMA3.1-8B & 98.08 & 17.03 & 52.64 \\
        Qwen2-VL~\cite{wang2024qwen2} & Qwen2-7B & 99.08 & 18.93 & 59.45\\
        MiniCPM-V 2.6~\cite{yao2024minicpm} & Qwen2-7B & 94.79 & 13.59 & 43.04 \\
        InternVL2~\cite{chen2024internvl} & InternLM2.5-7B & \underline{99.94} & 19.72 & 60.08 \\
        \hline
        SceneVLM & Qwen2-7B & \textbf{100} & \underline{67.85} & \underline{97.36} \\
        SceneVLM & InternLM2.5-7B & \textbf{100} & \textbf{74.32} & \textbf{97.42} \\
        \Xhline{2.5\arrayrulewidth}
    \end{tabular}
}
\caption{Accuracy of our method with other VLMs in distance estimation. ``Number'' indicates the percentage of responses that include numerical values. The best and the second results are highlighted in \textbf{blod} and \underline{underlined}, respectively.}
\label{tab:dis}
\end{table}

\subsection{Implementation}
\noindent
\textbf{Training details.}
We employ two advanced open-source VLMs, InternVL2~\cite{chen2024internvl} and Qwen2-VL~\cite{wang2024qwen2}, to train our sceneVLM. 
Specifically, both VLMs use their LLMs with 7B parameters, along with their open-source code.
We directly fine-tune these models on our SceneVQA dataset.

\noindent
\textbf{Evaluate metrics.}
For hierarchical scene graph generation, we convert the output JSON into pairwise relationship lists and evaluate the model's performance from four perspectives: Pairwise Relation Accuracy (PRA), Object-wise Relation Accuracy (OWA), Layer-wise Accuracy (LWA), and Node Detection Accuracy (NDA).
Each metric calculates its Precision, Recall, F-score, and Intersection over Union (IoU). 
For distance calculation tasks, we consider values between 50\%-200\% of the actual distance and refine the range to 80\%-120\% to enhance precision. 
For further details, please refer to the supplementary material.

\begin{figure*}[tbp]
\begin{center}
\includegraphics[width=1.0\linewidth]{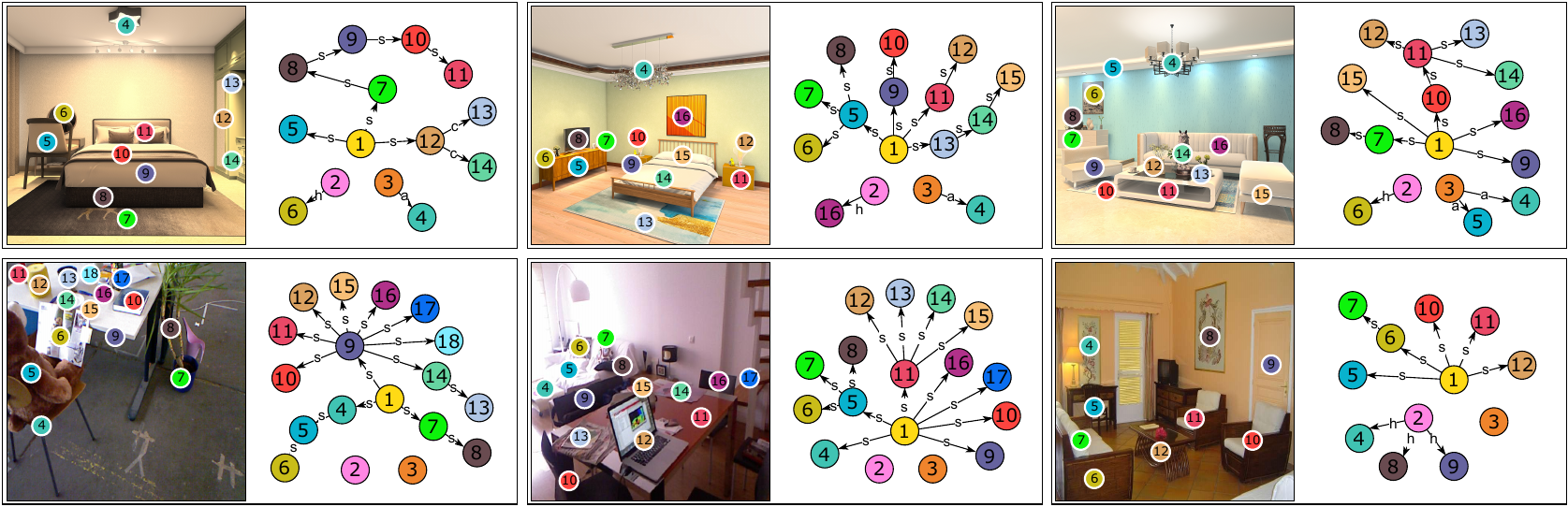}
\end{center}
\vspace{-0.1in}
\caption{Hierarchical scene graph visualization of our method. Each object is assigned a serial number, with the corresponding visual JSON is shown next to the image. Nodes represent objects, while edges indicate relationships. The numbers 1, 2, and 3 represent the floor, wall, and ceiling, respectively. For brevity, relationships such as ``support'', ``hang'', ``attach'', and ``contain'' are abbreviated to their initial letters. Object names are omitted from the labels to enhance clarity.}
\label{fig:qua_ssg}
\end{figure*}

\subsection{Scene Graph Generation}
As shown in Table~\ref{tab:sgg}, except for InstructBLIP~\cite{dai2023instruct} and LLaVA-1.5~\cite{liu2024improved}, JSON-formatted files can be successfully extracted from the results of most VLMs.
This capability is due to the inclusion of code data in the SFT dataset.
Analyzing from four distinct perspectives, the metrics for relationships (PRA and OWA) show minimal variation. 
In contrast, the metrics for objects (LWA and NDA) reveal significant discrepancies, attributable to LWA's stringent evaluation criteria, which require precise prediction at each node of the layers. 
Moreover, the metric for relationships is marginally lower than that for NDA, a disparity arising from the relative simplicity of object outputs compared to relationship outputs.
Given that the list of objects is provided, generating object outputs is relatively straightforward, whereas producing relationships is more complex, necessitating an understanding of indoor environments.
From the model perspective, our method outperforms existing VLMs across all metrics.
This improvement is attributed to the SceneVQA dataset, which facilitates scene graph generation for specific indoor scenes.
In terms of relationship metrics, both Precision and Recall are approximately 90\%, indicating a robust understanding of spatial relationships between indoor objects.
The evaluation metrics for object output show nearly 100\% accuracy, indicating that the model consistently outputs the entire provided object list without omissions.
Besides our method, GPT-4V is the next best-performing model achieving good results due to its strong generalization and comprehension capabilities.
However, other methods, despite accurately producing JSON-formatted files, tend to repeat examples from the question without fully understanding the instructional problem, leading to lower performance.

Moreover, Figure~\ref{fig:qua_ssg} visualizes the hierarchical JSON files produced by our method. 
The results demonstrate the model's effective comprehension of the depicted content and its ability to model the hierarchical relationships among objects within the room.

\subsection{Distance Estimation}
Most models are generally hesitant to provide numerical estimates when queried about spatial distances. 
To address this issue, we appended the instruction, ``Please output how many meters, for example: 2.1m,'' to the query. 
As shown in Table~\ref{tab:dis}, with the exception of InstructBLIP~\cite{dai2023instruct}, other models successfully predict distances rather than evading the question. 
Following the methodology of SpatialVLM~\cite{chen2024spatialvlm}, we assess the accuracy of the VLMs' predictions using a range defined by half to twice the ground truth distance.
Moreover, we narrow this range to [80,120] to enforce a more rigorous assessment, mirroring the typical use of approximate descriptions in daily life.
It is noteworthy that human descriptions of distances often exhibit imprecision, particularly when providing rough estimates. 
For example, a person might describe the length of a rope as approximately 1 meter instead of 1.12 meters.
Our method achieves the highest accuracy in both specified ranges, [80,120] and [50,200], surpassing other VLMs. 
In comparison, the performance of other VLMs in managing distance predictions is relatively inferior. 
This enhanced performance is primarily due to the SceneVLM dataset, which is rich in spatial information.
This enhancement allows our model to offer precise and reliable distance estimations, crucial for applications requiring accurate spatial understanding and measurements. 

Overall, the experimental results for both tasks validate the effectiveness of our sceneVLM in comprehending indoor environments, offering valuable new insights for advancing the field of indoor scene understanding.

\begin{table}[t]
\centering
\footnotesize
\renewcommand{\arraystretch}{1.2} 
\setlength{\tabcolsep}{10pt} 
\resizebox{0.9\linewidth}{!}{
    \begin{tabular}{lccc}
        \Xhline{2.5\arrayrulewidth}
         & Before iteration & After iteration & Change \\
        \hline
        Object count & 12.86 & 15.92 & +3.06 \\
        \hline
        Bbox area & 50472 & 18575 & -31897 \\
        New objects bbox area & - & 7943 & - \\
        \Xhline{2.5\arrayrulewidth}
    \end{tabular}
}
\caption{Average number of detected objects and bounding box areas before and after iteration, with an average image width of 1262 pixels and height of 1012 pixels.}
\label{tab:object_perception}
\end{table}

\begin{table*}[t]
\centering
\resizebox{\textwidth}{!}{%
\begin{tabular}{llllllllllllllllllllll}
    \Xhline{2.5\arrayrulewidth}
    \multirow{3}{*}{Method} & \multicolumn{4}{c}{Pairwise Relation Accuracy} & \multicolumn{4}{c}{Object-wise Relation Accuracy} & \multicolumn{4}{c}{Layer-wise Accuracy} & \multicolumn{4}{c}{Node Detection Accuracy} &  \multicolumn{2}{c}{Distance} \\
    \cmidrule(lr){2-5} \cmidrule(lr){6-9} \cmidrule(lr){10-13} \cmidrule(lr){14-17} \cmidrule(lr){18-19}
     & Precision & Recall & F-score & IoU & Precision & Recall & F-score & IoU & Precision & Recall & F-score & IoU & Precision & Recall & F-score & IoU & [80,120] & [50,200]  \\
    \hline
    Unforzen ViT &  90.3\textsubscript{\textcolor{mydarkblue}{-1.2}} & 89.7\textsubscript{\textcolor{mydarkblue}{-1.1}} & 89.9\textsubscript{\textcolor{mydarkblue}{-1.1}} & 83.8\textsubscript{\textcolor{mydarkblue}{-1.6}} & 86.8\textsubscript{\textcolor{mydarkblue}{-1.4}} & 86.3\textsubscript{\textcolor{mydarkblue}{-1.2}} & 86.5\textsubscript{\textcolor{mydarkblue}{-1.3}} & 79.0\textsubscript{\textcolor{mydarkblue}{-1.9}} & 76.4\textsubscript{\textcolor{mydarkblue}{-1.6}} & 76.5\textsubscript{\textcolor{mydarkblue}{-1.3}} & 76.2\textsubscript{\textcolor{mydarkblue}{-1.5}} & 70.4\textsubscript{\textcolor{mydarkblue}{-1.7}} & 99.8\textsubscript{\textcolor{mydarkgreen}{$\uparrow$0.1}} & 99.1\textsubscript{\textcolor{mydarkgreen}{$\uparrow$0.2}} & 99.4\textsubscript{\textcolor{mydarkgreen}{$\uparrow$0.1}} & 98.9\textsubscript{\textcolor{mydarkgreen}{$\uparrow$0.2}} & 77.8\textsubscript{\textcolor{mydarkgreen}{$\uparrow$3.5}} & 97.8\textsubscript{\textcolor{mydarkgreen}{$\uparrow$0.4}} \\
    w/o CoT &  80.9\textsubscript{\textcolor{mydarkblue}{-10.6}} & 81.0\textsubscript{\textcolor{mydarkblue}{-9.8}} & 80.8\textsubscript{\textcolor{mydarkblue}{-10.2}} & 76.8\textsubscript{\textcolor{mydarkblue}{-8.6}} & 78.8\textsubscript{\textcolor{mydarkblue}{-9.4}} & 78.6\textsubscript{\textcolor{mydarkblue}{-8.9}} & 78.7\textsubscript{\textcolor{mydarkblue}{-9.1}} & 73.6\textsubscript{\textcolor{mydarkblue}{-7.3}} & 71.8\textsubscript{\textcolor{mydarkblue}{-6.2}} & 71.6\textsubscript{\textcolor{mydarkblue}{-6.2}} & 71.5\textsubscript{\textcolor{mydarkblue}{-6.2}} & 67.6\textsubscript{\textcolor{mydarkblue}{-4.5}} & 87.0\textsubscript{\textcolor{mydarkblue}{-12.7}} & 86.8\textsubscript{\textcolor{mydarkblue}{-12.1}} & 86.9\textsubscript{\textcolor{mydarkblue}{-12.4}} & 86.5\textsubscript{\textcolor{mydarkblue}{-12.2}} & 73.6\textsubscript{\textcolor{mydarkblue}{-0.7}} & 97.3\textsubscript{\textcolor{mydarkblue}{-0.1}} \\
    w/o JSON &  91.0\textsubscript{\textcolor{mydarkblue}{-0.5}} & 90.3\textsubscript{\textcolor{mydarkblue}{-0.5}} & 90.5\textsubscript{\textcolor{mydarkblue}{-0.5}} & 84.9\textsubscript{\textcolor{mydarkblue}{-0.5}} & 87.4\textsubscript{\textcolor{mydarkblue}{-0.8}} & 86.9\textsubscript{\textcolor{mydarkblue}{-0.6}} & 87.1\textsubscript{\textcolor{mydarkblue}{-0.7}} & 79.9\textsubscript{\textcolor{mydarkblue}{-1.0}} & 77.3\textsubscript{\textcolor{mydarkblue}{-0.7}} & 77.1\textsubscript{\textcolor{mydarkblue}{-0.7}} & 77.0\textsubscript{\textcolor{mydarkblue}{-0.7}} & 71.4\textsubscript{\textcolor{mydarkblue}{-0.7}} & 99.5\textsubscript{\textcolor{mydarkblue}{-0.2}} & 98.8\textsubscript{\textcolor{mydarkblue}{-0.1}} & 99.1\textsubscript{\textcolor{mydarkblue}{-0.2}} & 98.6\textsubscript{\textcolor{mydarkblue}{-0.1}} & 73.0\textsubscript{\textcolor{mydarkblue}{-1.3}} & 97.1\textsubscript{\textcolor{mydarkblue}{-0.3}} \\
    Larger VLM &  93.2\textsubscript{\textcolor{mydarkgreen}{$\uparrow$1.7}} & 92.8\textsubscript{\textcolor{mydarkgreen}{$\uparrow$2.0}} & 92.9\textsubscript{\textcolor{mydarkgreen}{$\uparrow$1.9}} & 88.3\textsubscript{\textcolor{mydarkgreen}{$\uparrow$2.9}} & 90.4\textsubscript{\textcolor{mydarkgreen}{$\uparrow$2.2}} & 90.0\textsubscript{\textcolor{mydarkgreen}{$\uparrow$2.5}} & 90.2\textsubscript{\textcolor{mydarkgreen}{$\uparrow$2.4}} & 84.4\textsubscript{\textcolor{mydarkgreen}{$\uparrow$3.5}} & 82.1\textsubscript{\textcolor{mydarkgreen}{$\uparrow$4.1}} & 82.2\textsubscript{\textcolor{mydarkgreen}{$\uparrow$4.4}} & 81.9\textsubscript{\textcolor{mydarkgreen}{$\uparrow$4.2}} & 77.4\textsubscript{\textcolor{mydarkgreen}{$\uparrow$5.3}} & 100\textsubscript{\textcolor{mydarkgreen}{$\uparrow$0.3}} & 99.5\textsubscript{\textcolor{mydarkgreen}{$\uparrow$0.6}} & 99.7\textsubscript{\textcolor{mydarkgreen}{$\uparrow$0.4}} & 99.5\textsubscript{\textcolor{mydarkgreen}{$\uparrow$0.8}} & 82.3\textsubscript{\textcolor{mydarkgreen}{$\uparrow$8.0}} & 98.4\textsubscript{\textcolor{mydarkgreen}{$\uparrow$1.0}} \\
    \hline
    SceneVLM &  91.5 & 90.8 & 91.0 & 85.4 & 88.2 & 87.5 & 87.8 & 80.9 & 78.0 & 77.8 & 77.7 & 72.1 & 99.7 & 98.9 & 99.3 & 98.7 & 74.3 & 97.4 \\
    \Xhline{2.5\arrayrulewidth}
    \end{tabular}
    }
\caption{Ablation studies on various configurations of our SceneVLM.}
\label{tab:ablation}
\end{table*}

\subsection{Iterative Perception Performance}
The initial phase of iterative object perception in \rootname is crucial for the overall performance of the pipeline.
To evaluate this phase, we reassess the number of objects in our test dataset, as illustrated in Table~\ref{tab:object_perception}, which shows an average discrepancy of three objects between iterations.
Furthermore, we compute the mean area of object bounding boxes before and after iteration. 
The results demonstrate a decrease in area after iteration, enhancing the detection of smaller objects and facilitating a more precise interpretation of indoor environments.
Additionally, we calculate the area of object bounding boxes from the second iteration onwards. 
As shown in the last row of the table, objects detected in subsequent iterations are significantly smaller than those identified initially.
These metrics collectively demonstrate the significance of our iterative perception method.

\subsection{Ablative Experiments}
We conduct several ablation studies on SceneVLM to explore various aspects of the method, including unfreezing the vision encoder, the impact of CoT, using natural language to express relationships, and the effect of VLM size.
Experiments are performed on the InternVL2-8B~\cite{chen2024internvl}, and we use InternVL2-26B to validate the influence of model size.
The results of these ablation studies are shown in Table~\ref{tab:ablation}.

\noindent
\textbf{Unfrozen ViT.}
The results indicate that freezing or unfreezing the visual encoder does not significantly impact the outcomes. 
Unfreezing the encoder marginally reduces its performance in predicting object relationships, but slightly enhances its ability to estimate spatial distances. 
We infer that the visual encoder, extensively trained on a large SFT dataset, excels at modeling natural object relationships. 
Unfreezing it could slightly disrupt this established expertise. 
Considering its primary training involved extensive use of contrastive or classification losses, unfreezing it could lead to improved precision in fine-grained distance estimations, which is consistent with the observed results.

\noindent
\textbf{Impact of CoT.}
The influence of Chain-of-Thought (CoT) in predicting object relationships is pivotal. 
The results indicate that excluding CoT results in a decline in performance metrics by 5-10 points, and sometimes even more than 10 points.
This suggests that the model's direct outputs are less effective at handling complex relationship predictions. 
LLMs are particularly adept at extracting insights from natural language descriptions. 
For spatial relationship inference, the absence of CoT data causes performance to closely align with the established baseline.

\noindent
\textbf{Natural Language Relationship Output.}
In this study, we utilize JSON formatted outputs to depict relationships, which facilitates easy extraction of inter-object relationships. 
These relationships are represented in natural language as [subject, relation, object], where we observe a slight decline in performance across both tasks.

\noindent
\textbf{Different size of VLM.}
Intuitively, the size of the VLM significantly influences model performance. 
We explore this by scaling the VLM from 8B to 26B, observing enhancements across all performance metrics following expansion.
These improvements are consistent with our expectations.

\section{Applications beyond Understanding}

\begin{figure}[tbp]
\begin{center}
\includegraphics[width=1.0\linewidth]{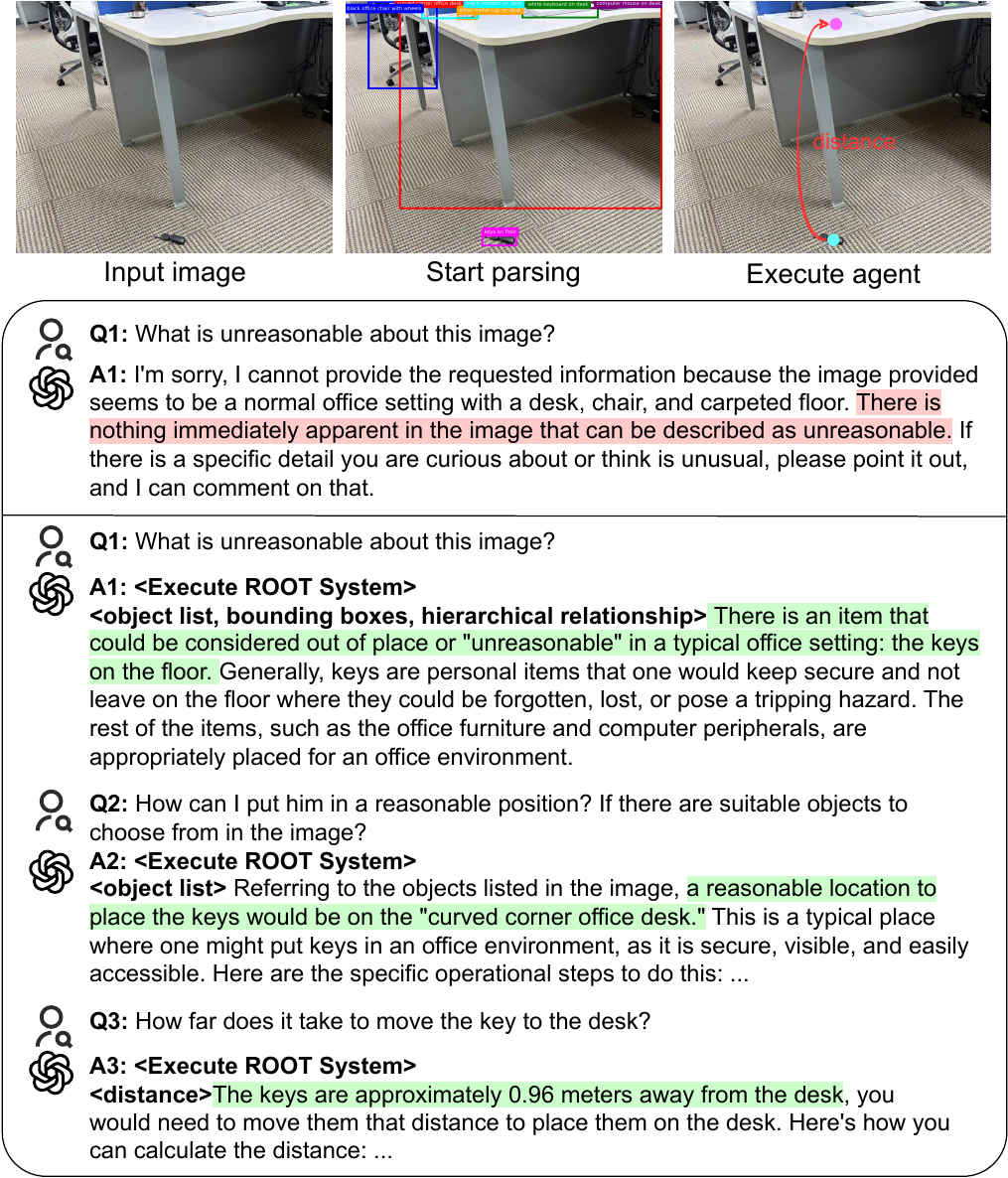}
\end{center}
\vspace{-0.1in}
\caption{\rootname and GPT-4V integration demonstrate a promising application in indoor environments. In an office setting, this system can assist a robot in accurately identifying and manipulating objects. By leveraging \rootname's analysis, GPT-4V identifies potential inconsistencies, thereby enhancing the coherence of indoor environments and reducing further economic losses.}
\label{fig:agent}
\end{figure}

We elucidate how a comprehensive understanding of indoor scenes can enhance downstream tasks. 
Here, we illustrate their utility in embodied AI applications, as well as their pivotal role in 3D scene generation.

\subsection{Embodied AI Integration}
As shown in Figure~\ref{fig:agent}, we explore a promising application of integrating the \rootname system with the GPT-4V model within the realm of embodied AI. 
The image is captured in a corner of a real office environment, simulating an indoor sweeping robot encountering a key on the floor. 
When directly queried about the discrepancies in the image, GPT-4V fails to provide an accurate response, potentially leading the robot to erroneously classify the key as trash.
After we employ the Root system to conduct a thorough analysis of the indoor scene.
\rootname assists GPT-4V in precisely identifying anomalies by recognizing objects and their hierarchical relationships.
Additionally, by incorporating the object list and leveraging the extensive world knowledge of a LLM, the robot can deduce the appropriate placement for the key.
Furthermore, with the \rootname providing spatial information between indoor objects, GPT-4V can deliver precise instructions to the robot, ensuring the key is placed correctly.
This integration not only improves the robot's scene comprehension but also prevents potential economic and safety risks due to inadequate comprehension.

\begin{figure}[tbp]
\begin{center}
\includegraphics[width=1.0\linewidth]{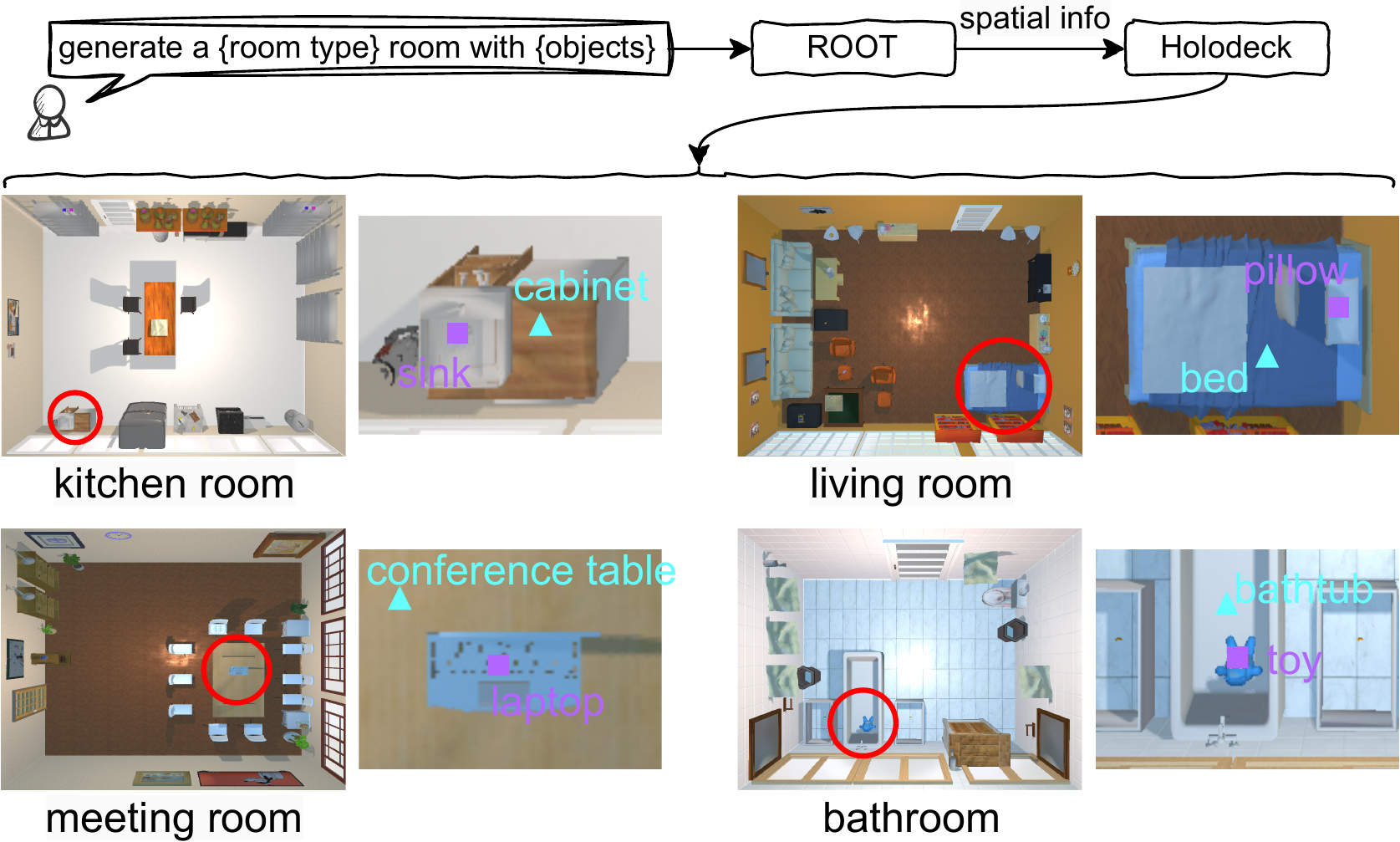}
\end{center}
\vspace{-0.1in}
\caption{\rootname's newly developed SceneLLM model, integrated with Holodeck~\cite{yang2024holodeck}, effectively constructs indoor scenes from specified object lists. We show several toy examples. For example, SceneLLM can define a structured relationship such as [cabinet, support, sink], allowing Holodeck to retrieve and assemble the corresponding objects based on these specifications.}
\vspace{-0.2in}
\label{fig:generation}
\end{figure}

\subsection{3D Scene Generation}
3D scene generation is pivotal in virtual reality environments and embodied AI simulations, where the scene authenticity significantly influences user experience and agent performance.
We propose that \rootname can improve the scene generation process.
This paper demonstrates that SceneVLM excels in generating hierarchical scene graphs from images, which accurately reflect the spatial arrangement of objects within a room.
We have retrained the process into SceneLLM, which now requires only a list of objects as input and utilizes its inherent knowledge to formulate the layout.
Subsequently, SceneLLM can organize the room layout based on user-specified objects.
This enhancement increases the flexibility and utility of scene generation in \rootname.

For instance, in kitchen design, users can specify various objects such as tables and bowls. 
SceneLLM constructs plausible hierarchical relationships among these objects and we utilize Holodeck~\cite{yang2024holodeck}, a language-guided system based on AI2-THOR~\cite{kolve2017ai2}, to integrate the SceneLLM model for managing object hierarchies and layouts.	
This integration streamlines the layout and modeling of object hierarchies within Holodeck.	
Users define the objects in the indoor environment, and SceneLLM generates their layouts.
Holodeck then retrieves objects with corresponding names from the objaverse~\cite{deitke2023objaverse}.
For horizontal structures, we continue to use Holodeck's methodology, applying constraints for horizontal optimization.	
Figure~\ref{fig:generation} displays scenes generated and optimized by Holodeck and SceneLLM.
The figure demonstrates that the optimized pipeline can generate indoor scenes with specified objects.
Users can provide extensive object information, enabling SceneLLM to tailor the generation of hierarchical relationships and enhance the realism and richness of the indoor scenes.
This method significantly enhances the flexibility and realism of scene generation.

\subsection{Border Impacts}
With the advancement in indoor scene understanding, we can now develop more complex applications such as layout arrangement, vision-language action models, and smart placement. 
Furthermore, this capability can be integrated into autonomous agents, enhancing their ability to perform intricate tasks traditionally handled by humans, such as household management.
This integration accelerates the advancement of automation and improves human convenience.
We believe that scene understanding is a critical component in achieving indoor Artificial General Intelligence (AGI), and the technological progress in this field has significantly propelled the evolution of our era.
Nonetheless, our methods exhibit certain limitations. 
For example, subsequent processes are heavily rely on the performance of the iterative object perception module, and our system struggles to achieve real-time scene analysis. 
We hope our method will inspire scholars and foster further advancements in this field.
\vspace{-0.1in}
\section{Conclusion}
In this paper, we introduce \rootname, a VLM-based system designed to comprehend indoor scenes by acquiring metadata of room objects and analyzing their spatial relationships.
Our experimental results reveal the limitations of current VLMs in interpreting indoor spaces and demonstrate the effectiveness of our approach. 
Additionally, we utilize the derived spatial information to enhance other applications, demonstrating its effectiveness.
We anticipate that \rootname will significantly impact the field of indoor scene understanding and inspire further research.


{
    \small
    \bibliographystyle{ieeenat_fullname}
    \bibliography{main}
}
\clearpage
\setcounter{page}{1}
\appendix
\maketitlesupplementary

\section{Details of Our \rootname System}
\subsection{Iterative Object Perception Algorithm}

We have detailed the execution process of our iterative object perception process in Algorithm~
\ref{alg:iterative_indoor_object_perception}.
In this algorithm, $I_{in}$ denotes the input image, and $gpt$ refers to the GPT-4V model, version dated 2024-07-01-preview. 
The term $dino$ represents GroundingDINO~\cite{liu2023grounding}. 
Additionally, we introduce several intermediate variables: 
$\{c_{i}\}_{i=1}^{N}$ indicates whether an object is a container, $\{p_{ij}\}_{i=1,j=1}^{N,M}$ quantifies the confidence level of the $j^{th}$ bounding box for the $i^{th}$ object, and $\{so_{i}\}_{i=1}^{N}$ signifies sub-objects.
Moreover, $\{sb_{ij}\}_{i=1,j=1}^{N,M}$ refers to the bounding boxes of sub-objects. 
We set a probability threshold of $p_{m}=0.3$ for exceeding specific criteria and set $p_{n}=0.15$ as the minimum probability required to discern the bounding box.
The scaling factor for iterative processes is denoted by $S=1.5$. 
The output includes $\{o_{i}\}_{i=1}^{N}$, indicating the objects, and $\{b_{ij}\}_{i=1,j=1}^{N,M}$, which are the bounding boxes associated with these objects.

\begin{figure*}[tbp]
\begin{center}
\includegraphics[width=0.98\linewidth]{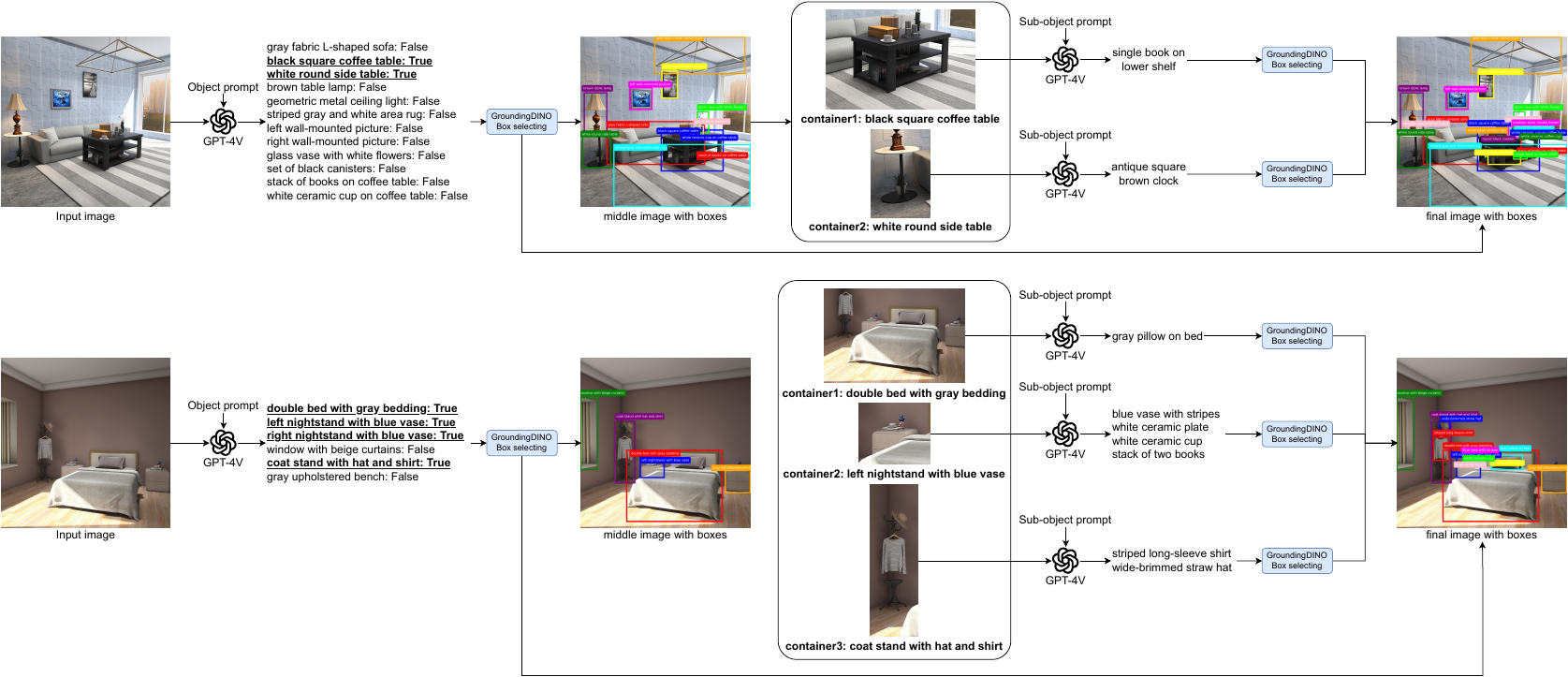}
\end{center}
\vspace{-0.1in}
\caption{The workflow for visualizing the iterative perception of objects.}
\label{fig:indoor_obj_perception}
\end{figure*}

Simultaneously, as shown in Figure~\ref{fig:indoor_obj_perception}, we visualize the entire execution process of the algorithm, thereby elucidating the workflow to ease the reader's comprehension.
This iterative approach is simple and effective for detecting small objects, such as books under a table or hats on a coat rack.

\begin{algorithm}[ht]
\caption{Iterative Object Perception}
\label{alg:iterative_indoor_object_perception}
\begin{algorithmic}[1]
\State \textbf{Input:} $I_{in}$
\State \textbf{Require:} $gpt$, $dino$
\State \textbf{Output:} $\{o_{i}\}_{i=1}^{N}$, $\{b_{ij}\}_{i=1,j=1}^{N,M}$
\State \textbf{Variables:} $\{c_i\}_{i=1}^N$, $\{p_{ij}\}_{i=1,j=1}^{N,M}$, $p_m$, $p_n$, $S$
\Function{FilterAndUpdate}{$\{b_{ij}\}, \{p_{ij}\}$}
    \State $\text{max\_p} \gets \max_{i,j} p_{ij}$
    \If{$\text{max\_p} > p_{m}$}
        \State $\{b'_{ij}\} \gets \{b_{ij} \mid p_{ij} \geq p_{m}\}$
        \State $\text{ps} \gets \text{sort descending}(\{p_{ij} \mid p_{ij} \geq p_{m}\})$
        \If{$\text{ps}[0]-\text{ps}[1] > p_n$}
            \State $\{b'_{ij}\} \gets \{b_{ij} \mid p_{ij} = \text{ps}[0]\}$
        \Else
            \State $\{b'_{ij}\} \gets gpt(\{b_{ij}\}, \text{``select prompt''})$
        \EndIf
    \Else
        \State \textbf{return} $\{\}$
    \EndIf
    \State \textbf{return} $\{b'_{ij}\}$
\EndFunction
\State \textbf{Start:}
\State $\{o_i\}, \{c_{i}\} \gets gpt(I_{\text{in}}, \text{``object prompt''})$
\State $\{b_{ij}\}, \{p_{ij}\} \gets dino(I_{\text{in}}, \{o_i\})$
\State $\{b_{ij}\} \gets \Call{FilterAndUpdate}{\{b_{ij}\}, \{p_{ij}\}}$
\State \textbf{Iterative Refinement for Containers:}
\For{$c_i = \text{True}$}
    \State $I_{\text{crop}} \gets \text{crop}(I_{\text{in}}, S \times b_{i})$ \Comment{Scale and crop image}
    \State $\{so_{i}\} \gets gpt(I_{\text{crop}}, \text{``sub-object prompt''})$
    \State $\{sb_{ij}\}, \{sp_{ij}\} \gets dino(I_{\text{crop}}, \{so_{i}\})$
    \State $\{sb_{ij}\} \gets \Call{FilterAndUpdate}{\{sb_{ij}\}, \{sp_{ij}\}}$
    \State Update $\{o_{i}\}$ and $\{b_{ij}\}$ with $\{so_{i}\}$ and $\{sb_{ij}\}$
\EndFor
\State \textbf{return} $\{o_i\}, \{b_{ij}\}$
\end{algorithmic}
\end{algorithm}

\subsection{Indoor Scene Parsing}
\begin{figure}[tbp]
\begin{center}
\includegraphics[width=1.0\linewidth]{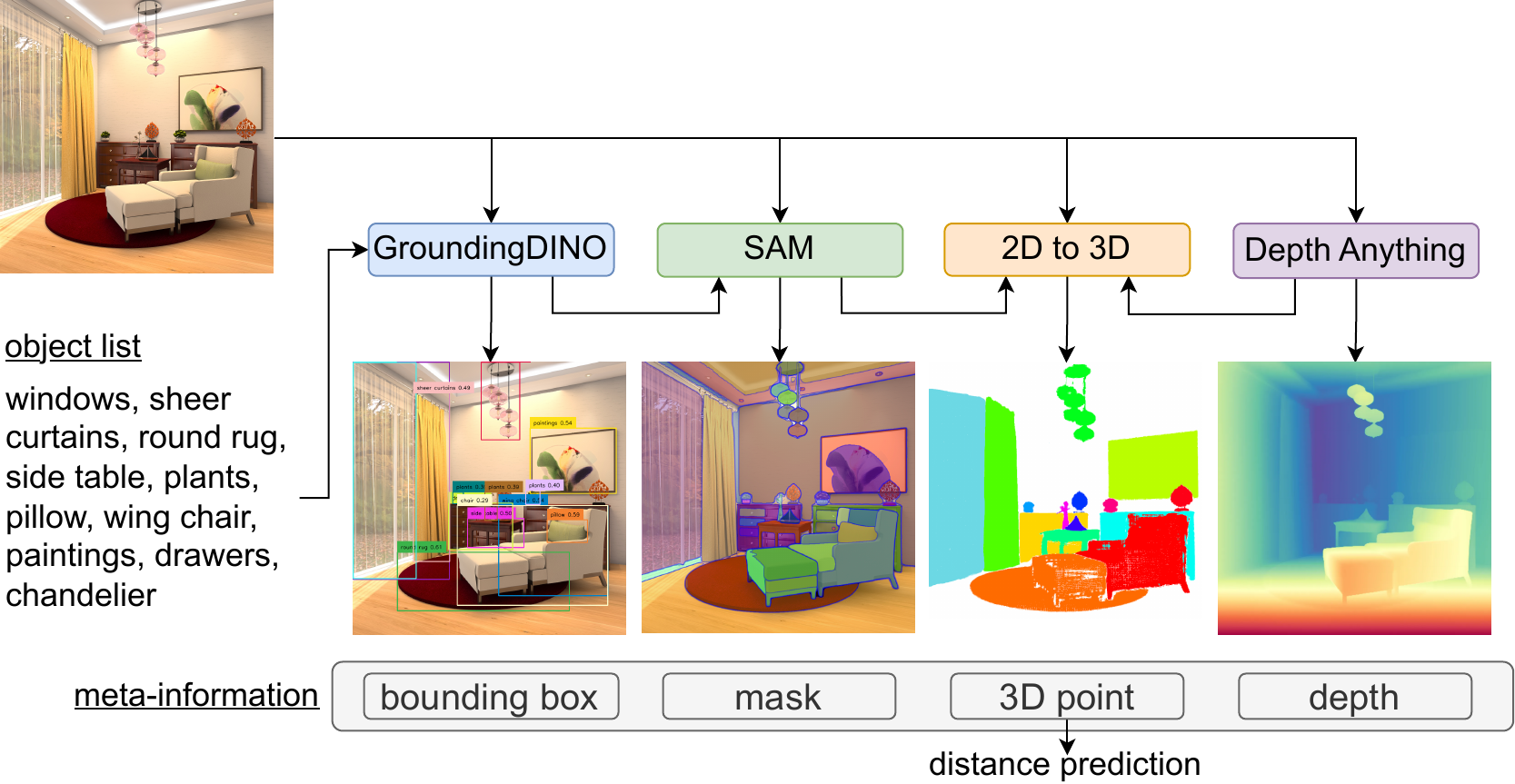}
\end{center}
\vspace{-0.1in}
\caption{The process of indoor scene parsing.}
\label{fig:indoor_scene_parsing}
\end{figure}

As shown in Figure~\ref{fig:indoor_scene_parsing}, the diagram details the process of acquiring additional meta-information, guided by the arrows.
During the iterative object detection phase, a list of objects along with their bounding boxes is generated.
The subsequent extraction of further meta-information leverages advanced vision foundation models, including SAM~\cite{kirillov2023segment} and DepthAnything~\cite{yang2024depth}.
 Following the indoor scene parsing process, we obtain a comprehensive list of objects within the scene, complete with their bounding boxes, masks, 3D points, and depth information.

\section{SceneVQA Dataset}
\subsection{Scene Data Collection}
Our scene dataset is collected from five sources: 3D-Future~\cite{fu20213d}, TUM~\cite{sturm2012benchmark}, SUN~\cite{xiao2016sun}, MIT Indoor Scenes~\cite{quattoni2009recognizing}, and Places~\cite{zhou2017places}. 
Since our focus is on indoor scenes, we exclude outdoor images and certain indoor images that do not meet our criteria. 
This includes close-ups of single objects, images with plain white backgrounds, and those depicting cartoons, sketches, or artwork.
We specifically concentrate on monocular indoor scenes.
To semantically filter the datasets, we employ the CLIP-ViT-H-14-378~\cite{fang2023data} model pre-trained on the DFN-5B dataset. 
Additionally, we refer to text prompts from SpatialVLM~\cite{chen2024spatialvlm} to define our positive and negative samples as follows:

\noindent
\textbf{Positive Samples:}
\begin{itemize}
    \item An iphone photo of an indoor scene.
\end{itemize}

\noindent
\textbf{Negative Samples:}
\begin{itemize}
    \item A close up shot of a single object.
    \item A product displayed in front of a white background.
    \item An artwork.
    \item A painting.
    \item A screenshot of graphics user interface.
    \item A piece of text.
    \item A sketch.
    \item A cartoon.
\end{itemize}

This approach ensures that the data used in our study is highly relevant and closely aligned with the specific requirements of our research on indoor scenes.

\subsection{Room Object Filtering}
In hierarchical relationships, it is generally understood that a room comprises three primary elements: floor, wall, and ceiling. 
These elements serve as the root nodes of the hierarchical scene graph. 
During the processing of 610,000 images through the \rootname pipeline, a total of 9,563,717 objects are identified. 
After eliminating duplicates, a refined list of 683,777 unique objects is established.
The objects are then filtered based on the following criteria:
\begin{itemize}
    \item Objects associated with the wall, ceiling, and floor, such as paneling.
    \item Objects exhibiting garbled data, a common issue with LLMs.
    \item Objects with non-English names.
    \item Objects not typically found indoors, such as mountains.
    \item Terms associated with humans, such as adult.
    \item Non-entity objects, such as window view.
\end{itemize}

Following these criteria, more than half of the objects are deemed unsuitable and are subsequently removed to enhance the quality of the remaining objects.
Ultimately, a curated list of 322,064 objects is retained and used to update our SceneVQA dataset.

\subsection{Room Types}

\begin{figure*}[tbp]
\begin{center}
\includegraphics[width=0.98\linewidth]{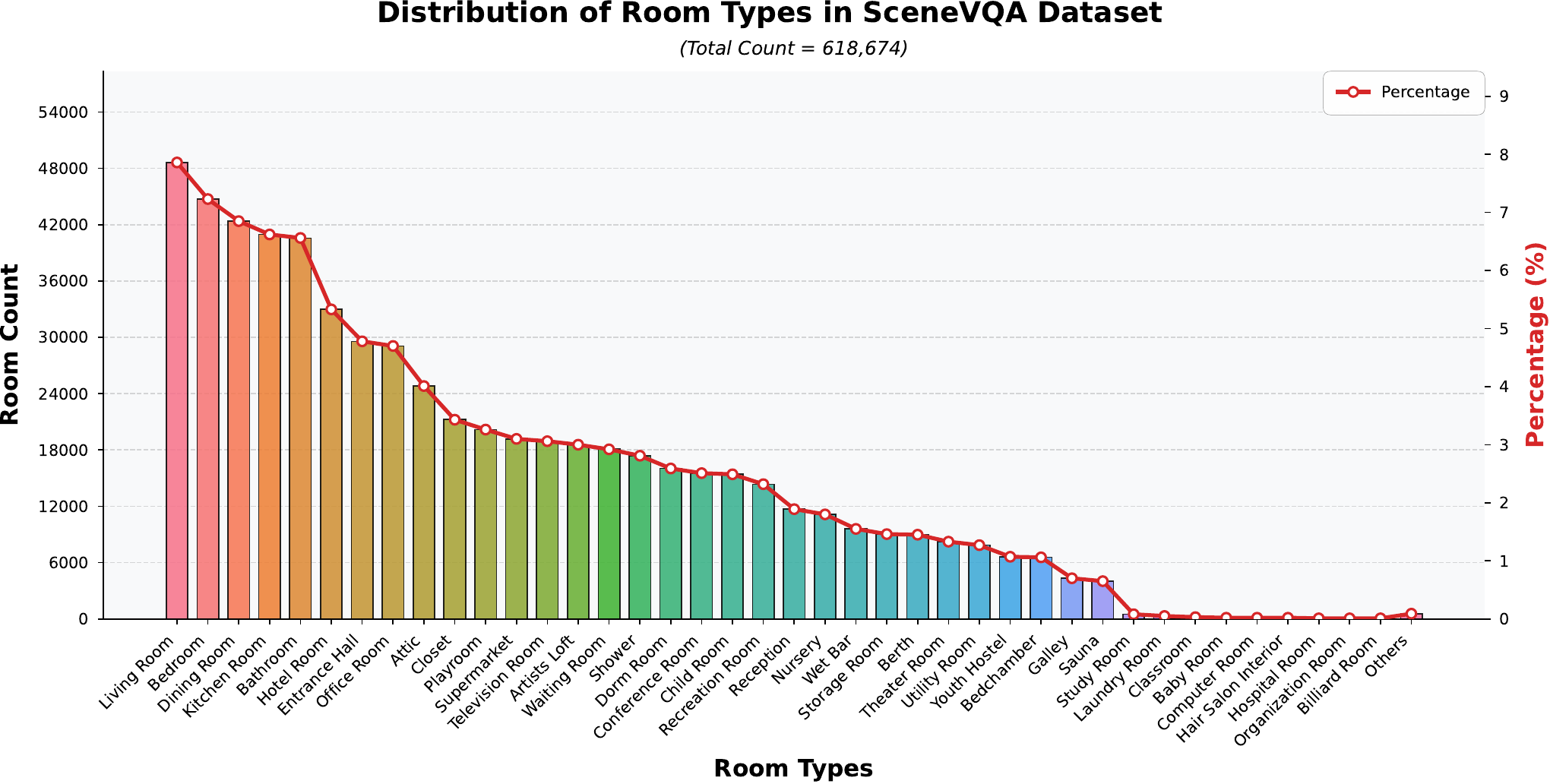}
\end{center}
\vspace{-0.1in}
\caption{Statistical distribution of room types in our SceneVQA dataset.}
\label{fig:sup_data_static}
\end{figure*}

Here, we have categorized the types of scenes.
As shown in Figure~\ref{fig:sup_data_static}, we have classified them into 41 categories (40 room types and an additional ``others'' category).
The classification of each scene is based on the metadata available in the dataset. 
For scenes without labels, we employ the GPT-4V to determine their final types. 
Additionally, we merge some categories that have similar meanings. 
The figure indicates that there are over 30 distinct scene types, each containing over 5000 images. 
Notably, prevalent indoor scenes such as living rooms and bedrooms each have over 40,000 images. 
This dataset has been employed to train our SceneVLM, improving its performance in novel scenes.

\section{Evaluation}

\begin{figure}[tbp]
\begin{center}
\includegraphics[width=1.0\linewidth]{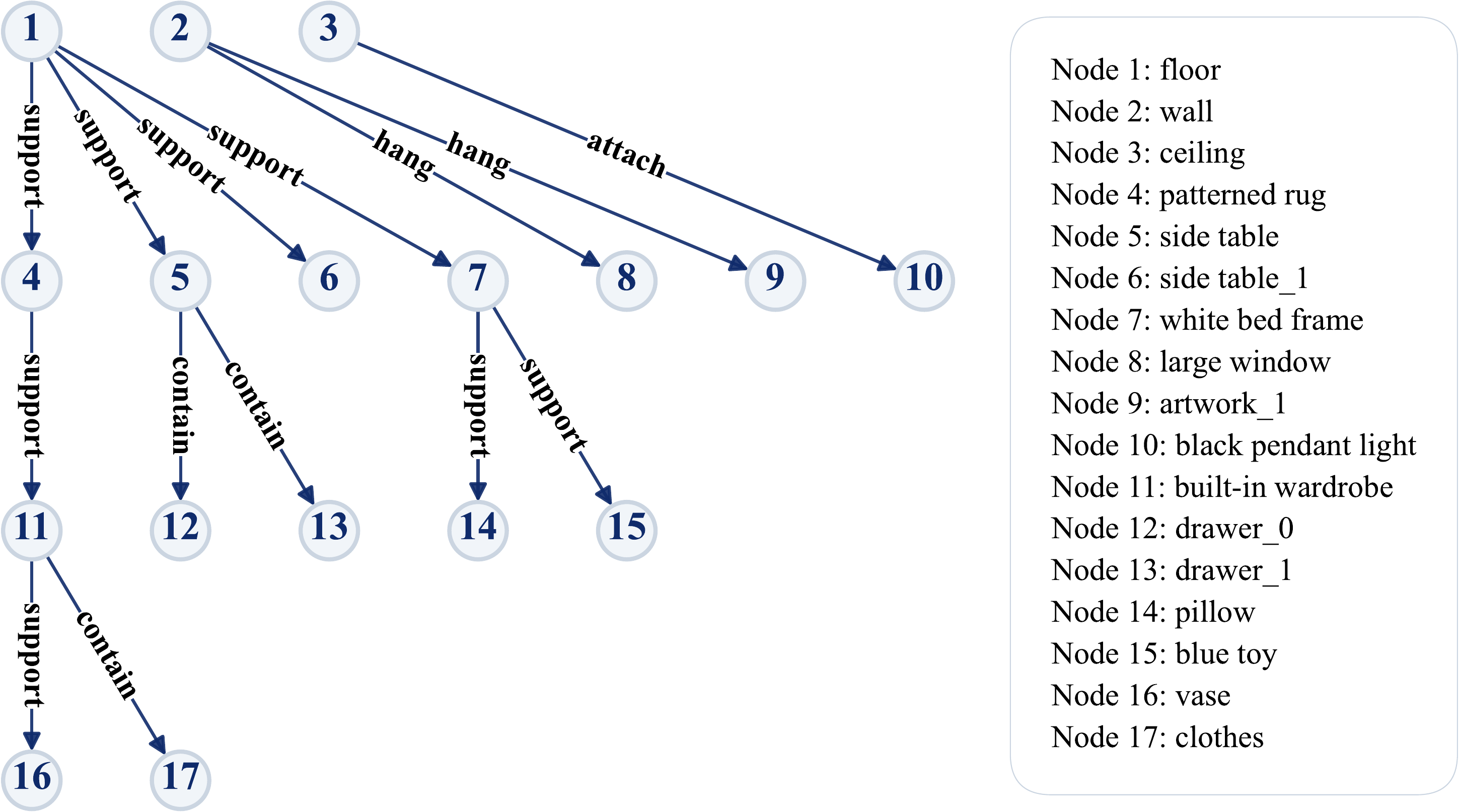}
\end{center}
\vspace{-0.1in}
\caption{An example of the JSON file representing hierarchical relationships for indoor objects.}
\label{fig:perspective}
\end{figure}

\subsection{Evaluation Perspectives}
The \rootname system outputs a JSON file that delineates hierarchical relationships among indoor objects.	
Traditional scene graph evaluation metrics may not be fully applicable in this context.
Consequently, we propose four perspectives to assess our method's performance, including Pairwise Relation Accuracy (PRA), Object-wise Relation Accuracy (OWA), Layer-wise Accuracy (LWA), and Node Detection Accuracy (NDA).
PRA and OWA represent the accuracy of the relationships between objects, while LWA and NDA represent the accuracy of the objects. 
As shown in Figure~\ref{fig:perspective}, we visualize a JSON file to exemplify these indicators.
Objects in the figure are labeled with serial numbers.

\noindent
\textbf{1. Pairwise Relation Accuracy (PRA):}
PRA assesses the accuracy of the relationships between pairs of objects.
For instance, in Figure~\ref{fig:perspective}, the relationship [1, support, 4] is considered correct if it is correctly extracted from the JSON file. 
There are 14 pairwise relationships in this example.	

\noindent
\textbf{2. Object-wise Relation Accuracy (OWA):}
OWA evaluates the accuracy of all relationships associated with a specific object, considering its parent and child objects.
For example, in Figure~\ref{fig:perspective}, object 1 has relationships such as [[1, support, 4], [1, support, 5], [1, support, 6], [1, support, 7]]. 
If all these relationships are accurately extracted from the JSON, the relationships for object 1 are considered precise. 
There are 17 object-wise relations in the figure.

\noindent
\textbf{3. Layer-wise Accuracy (LWA):}
LWA measures the accuracy of object predictions at each layer level. 
For example, in Figure~\ref{fig:perspective}, there are four layers: the first layer includes {1: {1,2,3}}, the second layer contains {2: {4,5,6,7,8,9,10}}, and so on.
Accuracy is achieved when both the layer level and all objects within that level are correctly predicted.

\noindent
\textbf{4. Node Detection Accuracy (NDA):}
NDA measures the accuracy of identifying individual objects.
In Figure~\ref{fig:perspective}, if an object, such as 1, appears in the JSON, it is considered as accurate. 
The figure contains a total of 17 objects.

For these metrics, we utilize precision, recall, F1-score, and Intersection over Union to derive quantitative results, which will be detailed in the following sections.

\subsection{Evaluation Metrics}

In this section, we compute four evaluation metrics: Precision, Recall, F1 Score, and Intersection over Union (IoU). 
These metrics are essential for assessing the performance of predictive models in our hierarchical scene graph generation.

Consider the following example, with ground truth (GT) = $\{a, b, c, d\}$ and predictions (pred) = $\{b, c, d, e, f\}$, we have:
\begin{itemize}
    \item True Positives (TP) = 3 (b, c, d)
    \item False Positives (FP) = 2 (e, f)
    \item False Negatives (FN) = 1 (a)
\end{itemize}
and we use this example to explain each metric.

\noindent
\textbf{1. Precision.} 
This metric quantifies the accuracy of the positive predictions made by the model. 
It is defined as the ratio of TP to the sum of TP and FP:

\begin{equation}
\text{Precision} = \frac{TP}{TP + FP} = \frac{3}{3 + 2} = \frac{3}{5} = 0.6
\end{equation}

\noindent
\textbf{2. Recall.}
This metric assesses the model's ability to identify all relevant instances.
It is defined as the ratio of TP to the sum of TP and FN:

\begin{equation}
\text{Recall} = \frac{3}{3 + 1} = \frac{3}{4} = 0.75
\end{equation}

\noindent
\textbf{3. F1 Score.}
This metric is the harmonic mean of Precision and Recall, providing a balanced measure of both metrics. 
It is computed as:

\begin{equation}
F1 = 2 \times \frac{\text{Precision} \times \text{Recall}}{\text{Precision} + \text{Recall}} = 2 \times \frac{0.6 \times 0.75}{0.6 + 0.75} = 0.67
\end{equation}

\noindent
\textbf{4. Intersection over Union (IoU).}
This metric evaluates the overlap between the predicted and ground truth sets. 
It is defined as the ratio of the area of overlap (TP) between the predicted and ground truth sets to the area of their union (TP + FP + FN):

\begin{equation}
\text{IoU} = \frac{TP}{TP + FP + FN} = \frac{3}{3 + 2 + 1} = \frac{3}{6} = 0.5
\end{equation}

\subsection{Evaluation Notes}

In this study, the test dataset consists of 740 images. 
However, the number of hierarchical relationships and distances far surpasses this count. 
Statistical analysis indicates that there are over 10,000 instances of relationships [subject, relation, object] between objects, and over 20,000 instances of distances in the form [object1, object2].

\section{Analysis of Distance Error}

\begin{figure}[tbp]
\begin{center}
\includegraphics[width=1.0\linewidth]{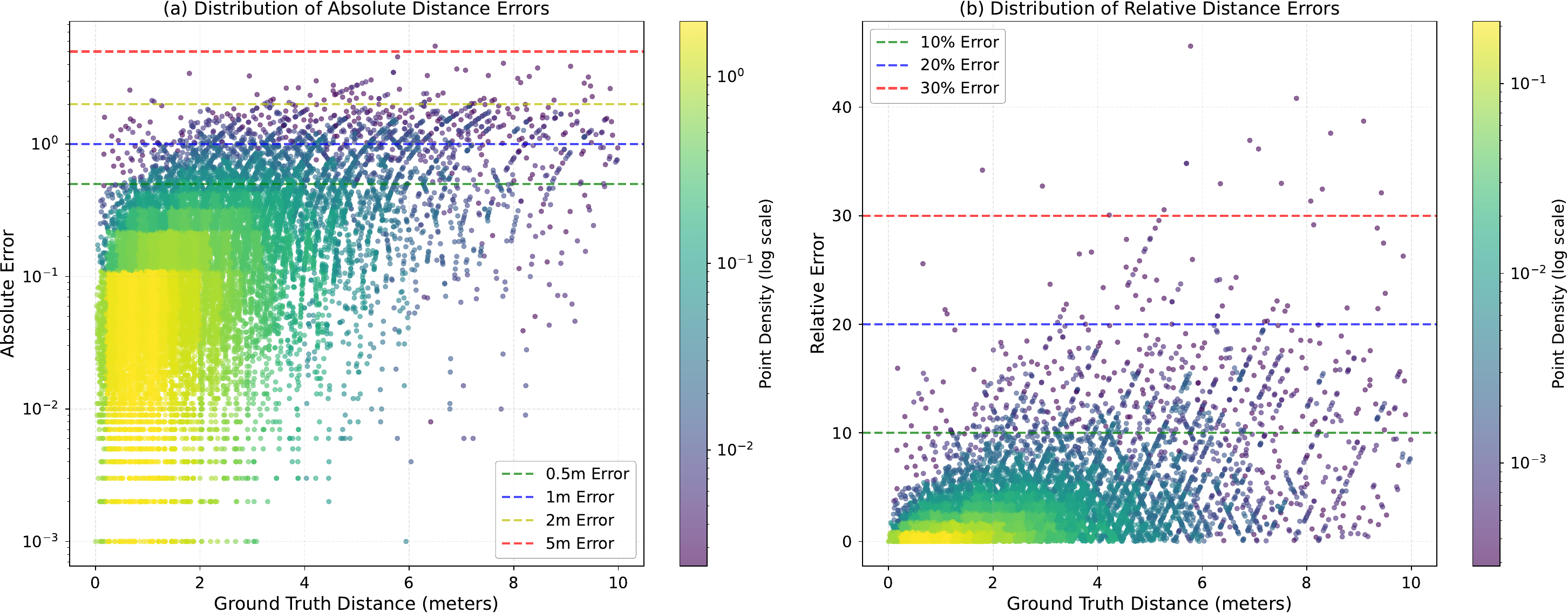}
\end{center}
\vspace{-0.1in}
\caption{Analysis of distance estimation errors: both absolute and relative errors escalate as the ground truth distance increases.}
\label{fig:sup_dis_error}
\end{figure}

Figure~\ref{fig:sup_dis_error} (left) illustrates the distribution of numerical errors, highlighting the error lines for distances of 0.5m, 1m, 2m, and 5m.
The majority of data points are located below the 2m error line, indicating that the prediction error for nearly all objects is less than 2m.
Additionally, a correlation is observed between smaller ground truth distances and smaller errors, with data points converging towards the 2m error line as the ground truth distance increases.
This trend indicates that the model exhibits enhanced performance in predicting proximal objects, likely due to features of nearby objects are more distinct and the DepthAnything model~\cite{yang2024depth} is more effective. 
In contrast, the increase in errors at larger distances can be attributed to reduced clarity of object features and a consequent loss of depth information, resulting in augmented noise.
In Figure~\ref{fig:sup_dis_error} (right), the distribution of relative errors is displayed. 
Here, the 10\%, 20\% and 30\% relative error lines are drawn.
As GT distances increase, points progressively deviate from the 0\% error line, reinforcing the observation from the left part of the figure that errors increase with larger GT distances.

\section{More Results in Holodeck}

\begin{figure*}[tbp]
\begin{center}
\includegraphics[width=0.98\linewidth]{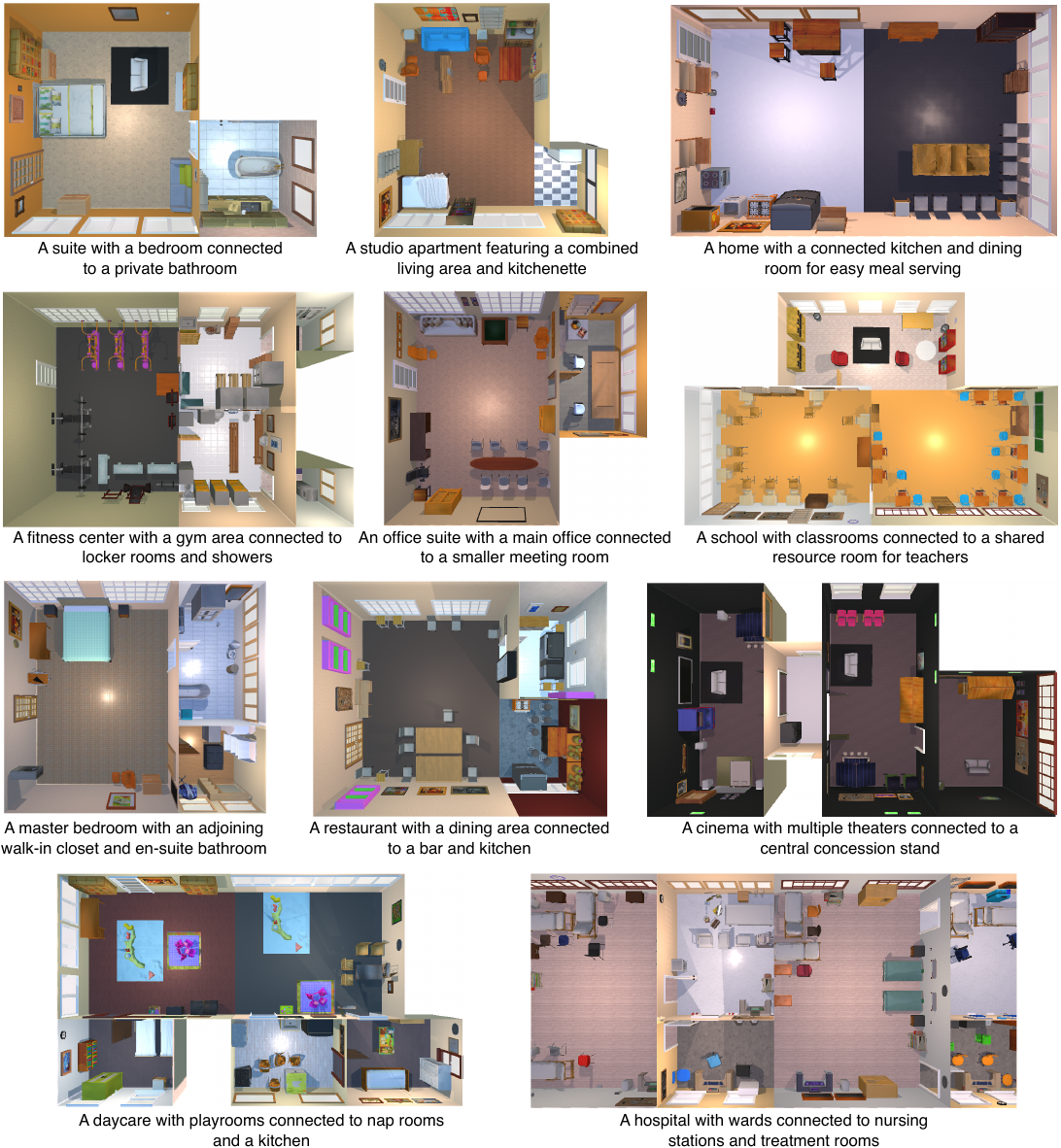}
\end{center}
\caption{More results on Holodeck~\cite{yang2024holodeck}. The integration of \rootname and Holodeck~\cite{yang2024holodeck} enhances functionality, enabling users to specify desired objects. Consequently, this integration facilitates the automation of indoor layout and arrangement processes.}
\label{fig:sup_holodeck}
\end{figure*}

In this section, we present additional results from integrating \rootname with Holodeck~\cite{yang2024holodeck}. 
Initially, we customize a collection of indoor objects and employ \rootname to define their hierarchical relationships. 
This hierarchy is then input into Holodeck, which utilized it as a basis to retrieve corresponding assets from Objaverse~\cite{deitke2023objaverse}. 
Holodeck disregard any objects that could not be retrieved and arrange the successfully retrieved objects in a horizontal layout based on predefined constraints.
Note that when the number of provided objects exceeds 20, the room might become congested depending on the room's size, the objects' sizes, and the placement rules. 
Consequently, some objects might be excluded from placement due to Holodeck's layout rules.
Moreover, users can opt to enlarge Holodeck's floorplan to accommodate more objects.
As shown in Figure~\ref{fig:sup_holodeck}, we exhibit the results of various indoor configurations. 
We supply distinct objects for different rooms. 
Users can also choose the objects they desire to place indoors, allowing \rootname and Holodeck to assist in the indoor planning process, thus simulating the entire indoor planning workflow.

\clearpage

\begin{figure*}[!t]
\begin{tcolorbox}[enhanced jigsaw, breakable, colback=black!5!white,colframe=black!75!black,title=Ojbect Perception Prompt: Obtain Object with Container]
\textbf{SYSTEM PROMPT: You are an assistant who perfectly describes images.}

Given an image, please create a JSON representation where each entry consists of a key ``object''  with a numerical suffix starting from 1. The value of each ``object'' key contains a ``description'' key and a ``container'' key, in which the value of the ``description'' key is a concise, up to eight-word sentence describing each main, clear, distinct object in the image while the ``container'' key's value should be either ``True'' or ``False'', indicating whether the targeted object has other sub-objects on or inside it.

Please note the following requirements:

1. Each entry should uniquely describe one element without repeating values. 

2. For the ``container'' key, its value should be ``True'' if the object is containing or supporting other objects, and ``False'' otherwise. 

3. The possible container that could only be a desk, shelf, bed or other similar items. Please consider a desk and its tablecloth as one object.

4. Do not miss any suitable object.

5. Ensure that your output can be parsed by python's  json.loads() directly.

Following is an example: \{``object1'': \{``description'': ``trash bin with liner'', ``container'': ``False''\}, ``object2'': \{``description'': ``retangular dinner table with tablecloths'', ``container'': ``True''\}, ``object3'': \{``description'': ``wooden shelf with electronic devices'', ``container'': ``True'' \}\}
\end{tcolorbox}
\end{figure*}

\begin{figure*}[!t]
\begin{tcolorbox}[enhanced jigsaw, breakable, colback=black!5!white,colframe=black!75!black,title=Ojbect Perception Prompt: Obtain Sub-object]
\textbf{SYSTEM PROMPT: You are an assistant who perfectly describes images.}

Given an image of a ``\{container\}'', please create a JSON representation where each entry consists of a key ``object'' with a numerical suffix starting from 1. The value of each ``object'' key contains a ``description'' key alue of the ``description'' key is a concise, up to eight-word sentence describing each main, clear, distinct object on or inside the ``\{container\}''.
Please note the following requirements:

1. Each entry should uniquely describe one element without repeating values. 

2. Only describe the objects that is on or inside the ``\{container\}''. Please ignore other parts of the image.

3. Do not miss any small object that is on or inside the ``\{container\}''.

4. Do not include the objects that are near, under or behind the ``\{container\}''. If there is no suitable object, please return -1.

5. Do not include the ``\{container\}'' in your output.

6. Ensure that the described objects are suitable for measuring distances between them and exclude elements like walls or floors. 

7. Make sure that your output can be parsed by python's  json.loads() directly.

Following is an example: \{``object1'': \{``description'': ``rectangular silver tray''\}, ``object2'': \{ ``description'': ``bottle of wine on table''\}, ``object3'': \{``description'': ``round decorative doily''\}\}
\end{tcolorbox}
\end{figure*}

\begin{figure*}[!t]
\begin{tcolorbox}[enhanced jigsaw, breakable, colback=black!5!white,colframe=black!75!black,title=Ojbect Perception Prompt: Select Bounding Boxes]
\textbf{SYSTEM PROMPT: You are an assistant who perfectly describes images.}

Please analyze an image that contains \{count\} bounding boxes. 
Each bounding box corresponds to one color. 
Your task is to identify the bounding box that best corresponds to the provided description of an object within the image and return the color of your selected bounding box. 

In the image, there are \{count\} bounding boxes.
The colors of these boxes include: \{colors\}.

Following is the requirement:

1. You must select the most appropriate bounding box and object based on orientation words within the description, such as ``left'', ``center/middle'' or ``right''. 
For instance, if an image contains three side-by-side computers, and the description states ``center computer'', you should output the color corresponding to the computer in the center.

2. It is possible that there are three similar objects (left, center and right respectively) in the image while only two of thems are enclosed by bounding boxes. 
In this situation, you still need to select the the suitable bounding box based on the relative position of these three objects.

3. Please provide an output in JSON format with the keys ``reason'' and ``color''. 
In the ``reason'' value, explain the rationale behind your selection, and in the ``color'' value, return the color of your chosen bounding box.

4. If there is no orientation word, you should select the bounding box that best corresponds to the given description. If none of the bounding box meets the description, you should select one randomly.

5. You can only select one box and the ``color'' value can only be one of the element from this color list: \{colors\}

6. The order of the color list is meaningless. 
You should select the bounding box and its corresponding color according to the description.

7. Make sure that your output can be parsed by python's json.loads() directly.

Following is the provided description: ``\{description\}'' 
\end{tcolorbox}
\end{figure*}

\begin{figure*}[!t]
\begin{tcolorbox}[enhanced jigsaw, breakable, colback=black!5!white,colframe=black!75!black,title=An toy example of an answer from GraphVQA: CoT and JSON]
The art frame is hanging on the wall. The \textnormal{bookshelf\_0}, desk, and chair are supported by the floor. On top of the desk, there are a mug, a toothbrush holder, and a notebook.

\begin{lstlisting}[]
{
    "wall": {
        "hang": [
            {
                "art frame": {}
            }
        ]
    },
    "ceiling": {},
    "floor": {
        "support": [
            {
                "bookshelf_0": {}
            },
            {
                "desk": {
                    "support": [
                        {
                            "mug": {}
                        },
                        {
                            "toothbrush holder": {}
                        },
                        {
                            "notebook": {}
                        }
                    ]
                }
            },
            {
                "chair": {}
            }
        ]
    }
}
\end{lstlisting}
\end{tcolorbox}
\end{figure*}

\begin{figure*}[!t]
\begin{tcolorbox}[enhanced jigsaw, breakable, colback=black!5!white,colframe=black!75!black,title=Prompt of GraphVQA]
Please determine the hierarchical relationships between the objects ({object list}) marked as point in the image. Use only these four hierarchical relationships: support, contain, attach, and hang.

For example, use ``support'' for objects on a table or chair, ``contain'' for objects inside a bookshelf or bottle, and ``hang'' for objects on the wall like doors, curtains, or paintings. Objects on the ceiling, such as lights, should use ``attach''. If there's a drawer in a table or objects inside the drawer, the relationship should be ``contain''. For objects on the floor, like tables on a carpet, the relationship is ``floor supports rug supports table''.

Present the relationships in a JSON tree format, with the ceiling, wall, floor as the root nodes. Here's an example JSON structure:
\begin{lstlisting}[]
{
    "ceiling": {
        "attach": [
            {
                "object": {}
            }
        ]
    }, 
    "wall": {}, 
    "floor": {
        "support": [
            {
                "object": {
                    "support": [
                        {
                            "object": {
                                "support": [
                                    {
                                        "object": {}
                                    }
                                ]
                            }
                        }, 
                        {
                            "object": {}
                        }
                    ]
                }
            },
            {
                "object": {}
            }
        ]
    }
}
\end{lstlisting}
\end{tcolorbox}
\end{figure*}

\begin{figure*}[!t]
\begin{tcolorbox}[enhanced jigsaw, breakable, colback=black!5!white,colframe=black!75!black,title=Prompt of DistanceVQA]
\textbf{Single Distance Queries:}
\begin{itemize}
    \item What's the distance from [A] to [B]?
    \item Can you calculate the length between [A] and [B]?
    \item Could you find out how far [A] is from [B]?
    \item Tell me how much space is between [A] and [B].
    \item Can you estimate the distance from [A] to [B]?
    \item What's the measurement of the distance between [A] and [B]?
    \item Do you know how many meters are between [A] and [B]?
    \item Can you tell the distance between [A] and [B]?
    \item How many steps would it take to get from [A] to [B]?
    \item Please measure the space between [A] and [B].
    \item How far would I need to walk to get from [A] to [B]?
    \item Please calculate the distance of [A] from [B].
    \item How many feet are between [A] and [B]?
    \item Could you provide an estimate of the distance from [A] to [B]?
    \item Can you measure how far [A] is from [B]?
\end{itemize}

\textbf{Dual Distance Queries:}
\begin{itemize}
    \item Can you determine the distance from [A] to [B] and also from [C] to [D]?
    \item What is the measurement of the space separating [A] and [B], and also [C] and [D]?
    \item Could you calculate the lengths between [A] and [B], and between [C] and [D]?
    \item Please provide the distances from [A] to [B] and from [C] to [D].
    \item How far apart are [A] and [B], and what about the distance between [C] and [D]?
    \item Can you estimate how many meters separate [A] from [B] and [C] from [D]?
    \item Tell me the distance between [A] and [B], and also calculate it for [C] and [D].
    \item Could you measure the space from [A] to [B] and compare it with the distance from [C] to [D]?
    \item What's the length from [A] to [B] and from [C] to [D]?
    \item How many steps would it take to walk from [A] to [B] and from [C] to [D]?
    \item Please estimate the distance between [A] and [B], and also between [C] and [D].
    \item Can you tell me how much space separates [A] from [B], and the same for [C] and [D]?
    \item How many feet are there between [A] and [B], and also between [C] and [D]?
    \item Could you inform me about the distances from [A] to [B] and from [C] to [D]?
    \item What are the measurements of the distances between [A] and [B], and [C] and [D]?
\end{itemize}

\textbf{Triple Distance Queries:}
\begin{itemize}
    \item Can you determine the distance from [A] to [B], and also from [C] to [D], and from [E] to [F]?
    \item Please calculate the lengths between [A] and [B], [C] and [D], and [E] and [F].
    \item How far is it from [A] to [B], and could you also tell me the distance between [C] and [D], and [E] and [F]?
    \item Could you measure the spaces between [A] and [B], [C] and [D], and [E] and [F]?
    \item What are the distances from [A] to [B], from [C] to [D], and from [E] to [F]?
    \item I need to know how many meters separate [A] and [B], [C] and [D], and [E] and [F]. Can you help?
    \item Can you provide the measurements of the distances between [A] and [B], [C] and [D], and [E] and [F]?
    \item How many steps would it take to walk from [A] to [B], from [C] to [D], and from [E] to [F]?
    \item Please inform me about the distance from [A] to [B], the distance from [C] to [D], and the distance from [E] to [F].
    \item Can you estimate how far [A] is from [B], how far [C] is from [D], and how far [E] is from [F]?
    \item What is the length from [A] to [B], from [C] to [D], and from [E] to [F]?
    \item Could you tell me how much space separates [A] and [B], [C] and [D], and [E] and [F]?
    \item How many feet are there between [A] and [B], between [C] and [D], and between [E] and [F]?
    \item Could you provide an estimate of the distances from [A] to [B], from [C] to [D], and from [E] to [F]?
    \item Please measure how far [A] is from [B], how far [C] is from [D], and how far [E] is from [F].
\end{itemize}
\end{tcolorbox}
\end{figure*}








\end{document}